\documentclass{article} 
\usepackage{iclr2025_conference,times}

\usepackage{tabularx}
\usepackage{makecell}
\usepackage{algpseudocode}
\usepackage{booktabs}
\usepackage[export]{adjustbox}
\usepackage{hhline}
\usepackage{xspace}
\usepackage{xcolor}
\usepackage{colortbl} 
\usepackage{pifont}
\usepackage{graphicx}
\usepackage{wrapfig}
\usepackage{trajan}
\usepackage{capt-of}
\usepackage{multicol}
\usepackage{multirow}
\usepackage{inconsolata}
\usepackage{listings}
\usepackage{longtable}
\usepackage{array}
\usepackage{todonotes}
\usepackage[cachedir=.,frozencache]{minted}

\newcolumntype{L}[1]{>{\raggedright\arraybackslash}p{#1}} 
\newcolumntype{C}[1]{>{\centering\arraybackslash}p{#1}}   

\usepackage{verbatim}
\usepackage[labelformat=simple]{subcaption}

\definecolor{Box1Color}{RGB}{227, 236, 246}
\definecolor{Box2Color}{RGB}{248, 220, 225}
\definecolor{Box3Color}{RGB}{255, 238, 224}

\newcommand{\sharegpto}{\textsc{ShareGPT-4o-Reasoning}\xspace}

\newcommand{\llavanext}{\textsc{LLaMA3-LLaVA-NeXT-8B}\xspace}
\newcommand{\openllavanext}{\textsc{Open-LLaVA-NeXT}\xspace}
\newcommand{\modelbase}{\textsc{LLaVA-NeXT-8B}\xspace}
\newcommand{\modelformat}{\textsc{LLaVA-NeXT-Format}\xspace}
\newcommand{\modeldirect}{\textsc{LLaVA-NeXT-Direct}\xspace}

\newcommand{\modelsft}{\textsc{LLaVA-Reasoner-SFT}\xspace}
\newcommand{\modelrl}{\textsc{LLaVA-Reasoner-DPO}\xspace}

\newcommand{\cOne}{\ding{172}\xspace}
\newcommand{\cTwo}{\ding{173}\xspace}
\newcommand{\cThree}{\ding{174}\xspace}
\newcommand{\cFour}{\ding{175}\xspace}
\newcommand{\cFive}{\ding{176}\xspace}
\newcommand{\cSix}{\ding{177}\xspace}

\newcommand{\aokvqa}{A-OKVQA\xspace}
\newcommand{\aok}{A-OK\xspace}
\newcommand{\chartqa}{ChartQA\xspace}
\newcommand{\docvqa}{DocVQA\xspace}
\newcommand{\infovqa}{InfoVQA\xspace}
\newcommand{\textvqa}{TextVQA\xspace}
\newcommand{\ocrbench}{OCRBench\xspace}
\newcommand{\mathvista}{MathVista\xspace}
\newcommand{\scienceqa}{ScienceQA\xspace}
\newcommand{\sqa}{SQA\xspace}
\newcommand{\aitd}{AI2D\xspace}
\newcommand{\mmmu}{MMMU\xspace}
\newcommand{\mmstar}{MMStar\xspace}
\newcommand{\mathvision}{MathVision\xspace}
\newcommand{\gllava}{G-LLaVA\xspace}

\newcommand{\gpto}{{GPT-4o}\xspace}
\newcommand{\chatgpt}{{ChatGPT}\xspace}

\newcommand{\highc}[1]{\textcolor{orange}{#1}}
\newcommand{\highd}[1]{\textcolor{blue}{#1}}

\usepackage{times}
\usepackage{latexsym}

\usepackage[T1]{fontenc}

\usepackage[utf8]{inputenc}

\usepackage{microtype}

\usepackage{inconsolata}

\usepackage{graphicx}

\usepackage{amsmath,amsfonts,bm}

\def\eqref#1{equation~\ref{#1}}

\def\1{\bm{1}}

\DeclareMathAlphabet{\mathsfit}{\encodingdefault}{\sfdefault}{m}{sl}
\SetMathAlphabet{\mathsfit}{bold}{\encodingdefault}{\sfdefault}{bx}{n}

\newcommand{\sigmoid}{\sigma}

\usepackage{hyperref}
\usepackage{url}
\usepackage{cleveref}
\usepackage{fontawesome5}

\title{Improve Vision Language Model Chain-of-thought Reasoning}

\newcommand{\github}{\raisebox{-0.13em}\faGithub}

\newcommand{\cmu}{\spadesuit}

\author{Ruohong Zhang$^{\cmu}$, Bowen Zhang$^{\dag}$, Yanghao Li$^{\dag}$, Haotian Zhang$^{\dag}$, Zhiqing Sun$^{\cmu}$, \\
\textbf{Zhe Gan$^{\dag}$, Yinfei Yang$^{\dag}$, Ruoming Pang$^{\dag}$, Yiming Yang$^{\cmu}$} \\
$^\cmu$CMU LTI, $^{\dag}$ Apple \\
\hspace{0.2\linewidth}\github~\texttt{\href{https://github.com/RifleZhang/LLaVA-Reasoner-DPO}{github.com/RifleZhang/LLaVA-Reasoner-DPO}}
}

\iclrfinalcopy 
\begin{document}

\maketitle

\begin{abstract}
Chain-of-thought (CoT) reasoning in vision language models (VLMs) is crucial for improving interpretability and trustworthiness. However, current training recipes lack robust CoT reasoning data, relying on datasets dominated by short annotations with minimal rationales. In this work, we show that training VLM on short answers does not generalize well to reasoning tasks that require more detailed responses. To address this, we propose a two-fold approach. First, we distill rationales from \gpto model to enrich the training data and fine-tune VLMs, boosting their CoT performance. Second, we apply reinforcement learning to further calibrate reasoning quality. Specifically, we construct positive (correct) and negative (incorrect) pairs of model-generated reasoning chains, by comparing their predictions with annotated short answers. Using this pairwise data, we apply the Direct Preference Optimization algorithm to refine the model's reasoning abilities. Our experiments demonstrate significant improvements in CoT reasoning on benchmark datasets and better generalization to direct answer prediction as well. This work emphasizes the importance of incorporating detailed rationales in training and leveraging reinforcement learning to strengthen the reasoning capabilities of VLMs.

\end{abstract}

\section{Introduction}

Chain-of-thought (CoT) reasoning is essential for improving the interpretability and trustworthiness of VLMs~\citep{onevision, liu2024llavanext, chen2023minigpt,liu2023visual, liu2023improved, bai2023qwen}. As VLMs are increasingly applied to more difficult tasks, the ability to reason through complex problems becomes essential. However, current training approaches for VLMs often rely on datasets dominated by short answers with limited rationales, which may restrict the models’ ability to generalize to tasks with comprehensive reasoning. In this work, we aim to address these limitations by providing distilled CoT data, introducing supervised finetuning (SFT) and reinforcement learning (RL) strategies to improve VLM reasoning performance.

\begin{figure}
    \centering
    \includegraphics[width=0.9\linewidth]{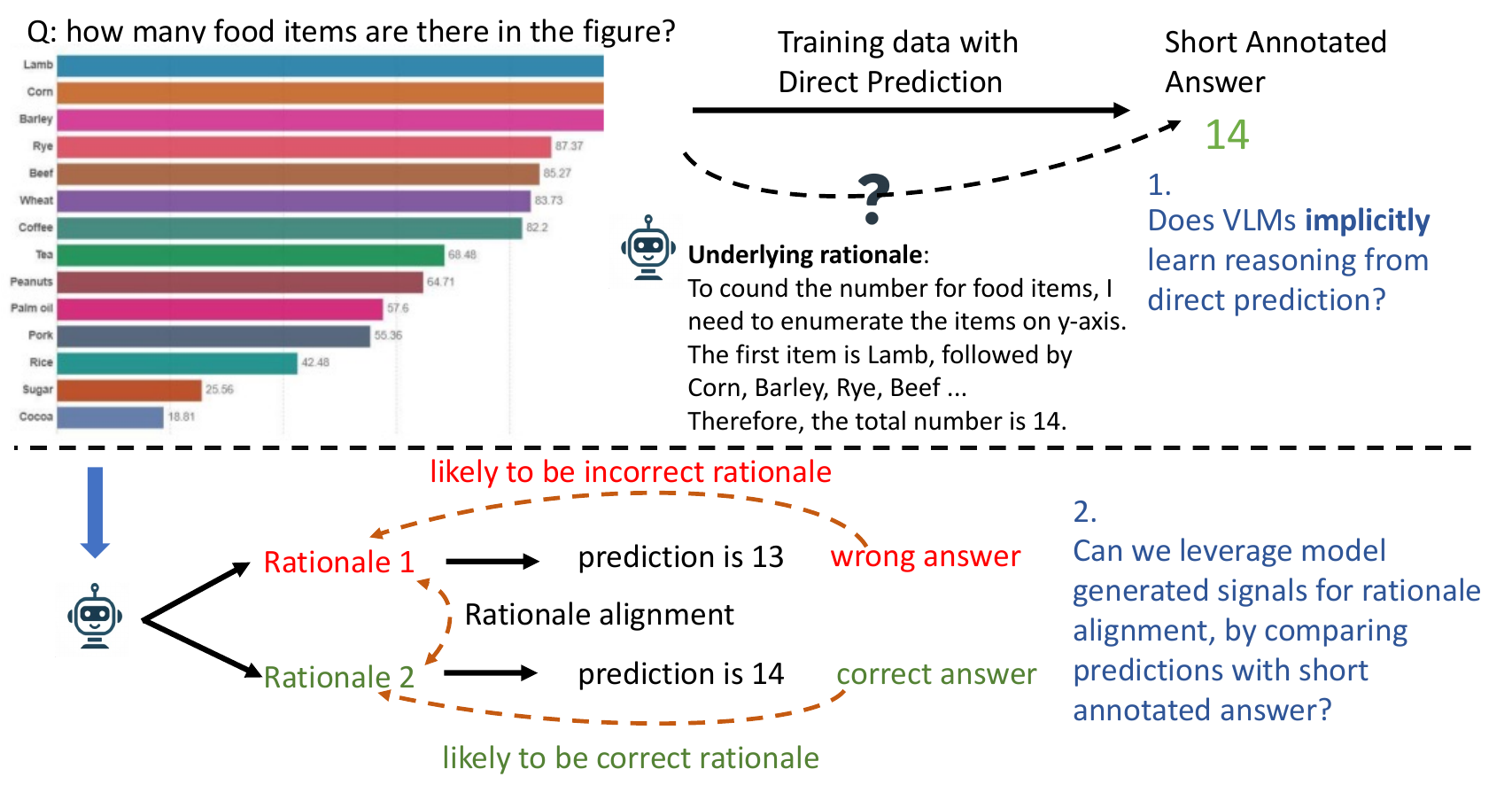}
    \caption{\footnotesize The upper figure questions whether training exclusively on direct-answer datasets can effectively teach CoT prediction. In the lower figure, generating CoT for prediction provides the additional benefit of reasoning alignment, allowing the model to improve by leveraging self-generated data.\label{fig:intro}}
\end{figure}

An example in \cref{fig:intro} asks for the number of food items in a bar graph. When answering this question, a human would typically enumerate the bars and then calculate the total. However, writing out this enumeration process is far more cumbersome than simply providing the short answer of “14.” Consequently, the annotated training data is predominantly composed of short answers, with minimal rationale provided. This raises a critical research question: \textit{Does training on direct prediction implicitly teach the model to perform chain-of-thought reasoning to derive correct answers?} Our findings indicate that after training on 26k direct predictions from \chartqa, the accuracy of direct predictions increased by 2.9 (70.2 to 73.1), while CoT prediction accuracy improved by only 0.6 points (71.2 to 71.8), with CoT under-performing direct prediction as a result. This suggests that current training approaches have limited effectiveness in enhancing CoT reasoning.


We hypothesize that developing CoT reasoning capabilities requires explicit training on data that includes detailed reasoning steps. To address the scarcity of high quality CoT reasoning data, we propose leveraging datasets with short ground truth annotations and employing the \gpto model to generate reasoning paths that lead to the correct answer. Our approach encompasses a diverse range of tasks, utilizing 9 datasets that demand different reasoning skills, including common world knowledge (\aokvqa), chart interpretation (\chartqa), document information localization (\docvqa, \infovqa), real-world text extraction (\textvqa), scientific reasoning (\aitd, \sqa), and mathematical reasoning (\mathvision, \gllava). We distilled a total of 193k CoT examples for supervised fine-tuning (SFT) and the model, \modelsft, demonstrates significant improvements in VLM chain-of-thought reasoning performance.

In the lower part of \cref{fig:intro}, we propose further refining SFT model reasoning through model-generated signals~\citep{sun2024easy, setlur2024rl}. Specifically, the model generates multiple CoT steps to derive final predictions, which are then compared to the provided short annotation. Rationales that lead to correct predictions are more likely to be accurate, and vice versa. By optimizing positive (correct) and negative (incorrect) pairs of rationales with Direct Preference Optimization (DPO), we align the VLM toward more accurate reasoning process. 
The aligned model, \modelrl, shows improved performance across all three domains as well as better out-of-domain generalization. Additionally, we demonstrate that DPO model can serve as a verifier to assign appropriate rewards, facilitating effective credit assignment~\citep{rafailov2024r, lu2024step}.

Our contributions can be summarized as follows:
(A) We release a comprehensive CoT dataset \sharegpto containing 193k examples, covering various VQA tasks.
(B) We demonstrate the effectiveness of SFT in improving CoT reasoning using this dataset.
(C) We show that reinforcement learning with DPO can further improve model reasoning using model-generated signals, without requiring additional human-labeled data.

\vspace{-0.3cm}

\section{Related Work}
\vspace{-0.3cm}

\paragraph{VLM Reasoning}
Previous work has evaluated the reasoning capabilities of vision-language models (VLMs) across various domains, including mathematics~\citep{mathvista, mathvision}, college-level questions~\citep{mmmu}, and science~\citep{ai2d, scienceqa}. Studies such as \cite{mavis, zhang2024tinychart, gllava} focus on training VLMs to generate step-by-step solutions for math problems or chart-based calculations, while \cite{shao2024visual} trains VLMs for chain-of-thought (CoT) reasoning in object localization tasks.

\vspace{-0.3cm}
\paragraph{VLM/LLM Alignment}
VLM alignment has utilized preference modeling techniques, such as Direct Preference Optimization (DPO)\citep{ouali2024clip, deng2024enhancing, yu2024rlaif, li2023silkie, gunjal2023detecting, sun2023aligning}, and Proximal Policy Optimization (PPO)\citep{sun2023aligning}, to improve factual accuracy and reduce hallucination. To improve reasoning capabilties in LLM, \cite{sun2024easy,setlur2024rl,lu2024step,pang2024iterative,xie2024monte} use iterative or step DPO to improve math CoT reasoning capabilities.
\begin{figure*}[ht]
    \centering
    \includegraphics[width=0.9\linewidth]{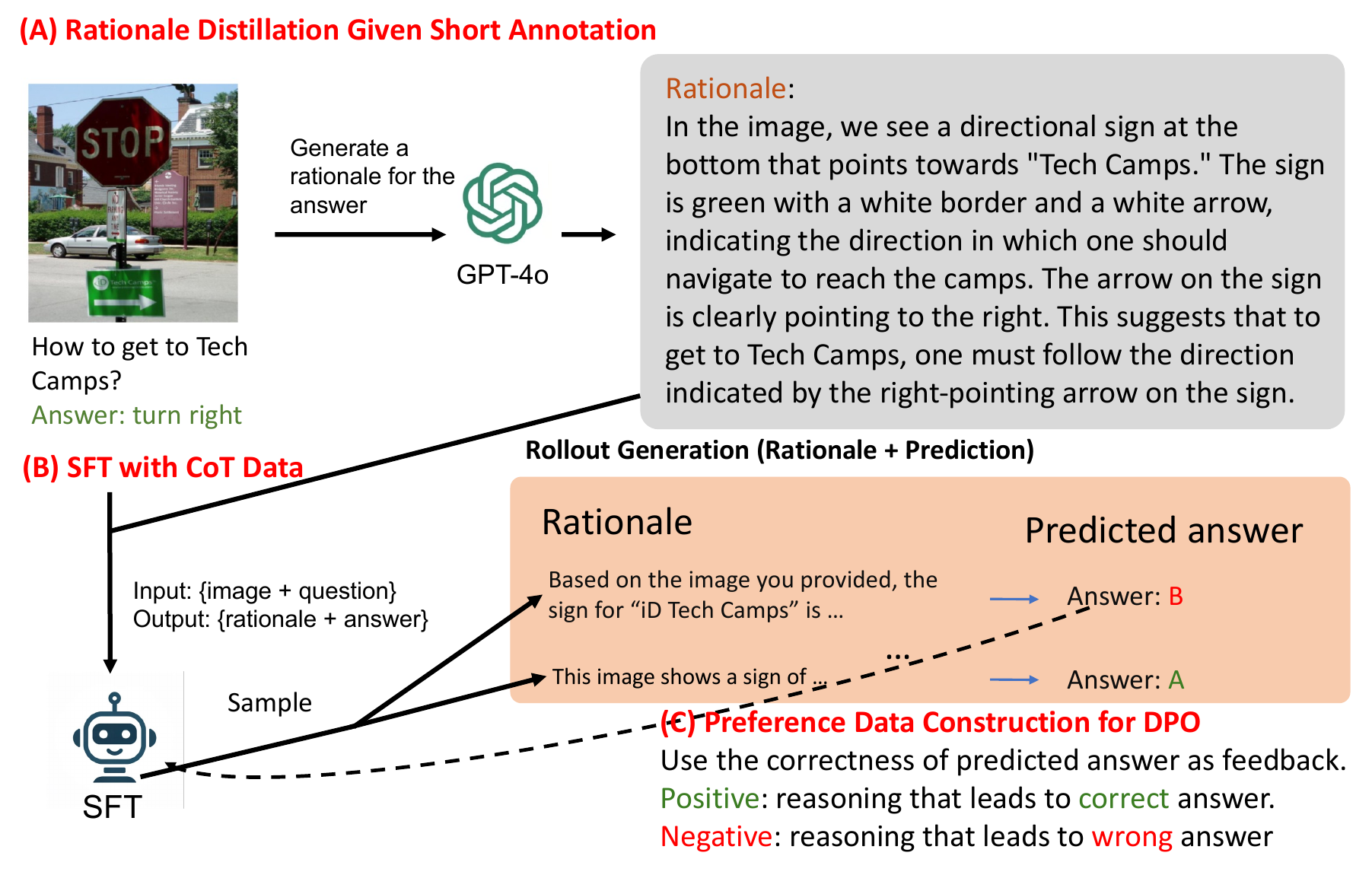}
    \caption{Workflow diagram showing: a) the use of \gpto to generate rationale given short annotations; b) SFT of open-source VLM for CoT reasoning; c) Build preference dataset for reinforcement learning with DPO to enhance reasoning.}
    \label{fig:illustration}
\end{figure*}

\begin{wraptable}{r}{0.3\linewidth}
\small
\vspace{-15pt}
\caption{Data statistics of CoT distilled on different dataset \label{tab:cot_data_stats}
}
\begin{tabularx}{\linewidth}{lc}
\hline
\textbf{Dataset} & \textbf{Dataset Size} \\ \hline
\aokvqa & 16.9k \\ 
\chartqa & 26.0k \\
\sqa & 6.1k \\
\aitd & 11.9k \\
\infovqa & 22.4k \\
\docvqa & 37.3k \\
\textvqa & 29.7k \\
\mathvision & 11.0k \\
\gllava & 30.3k \\ \hline
\textbf{Total} & 193k
\end{tabularx}
\vspace{-40pt}
\end{wraptable}


\section{Method}
\label{sec:method}
As shown in \cref{fig:illustration}, our pipeline consists of three stages: (A) CoT data distillation from \gpto (\cref{method:distill}), (B) SFT with CoT (and direct) data to enable VLM CoT reasoning, and (C) RL for further enhancement of CoT reasoning. The RL stage involves generating positive (correct) and negative (incorrect) reasoning data pairs sampled from SFT, as detailed in \cref{method:rl}.

\subsection{Reasoning Data Distillation}
\label{method:distill}
\begin{figure*}[ht]
    \centering
    \includegraphics[width=\linewidth]{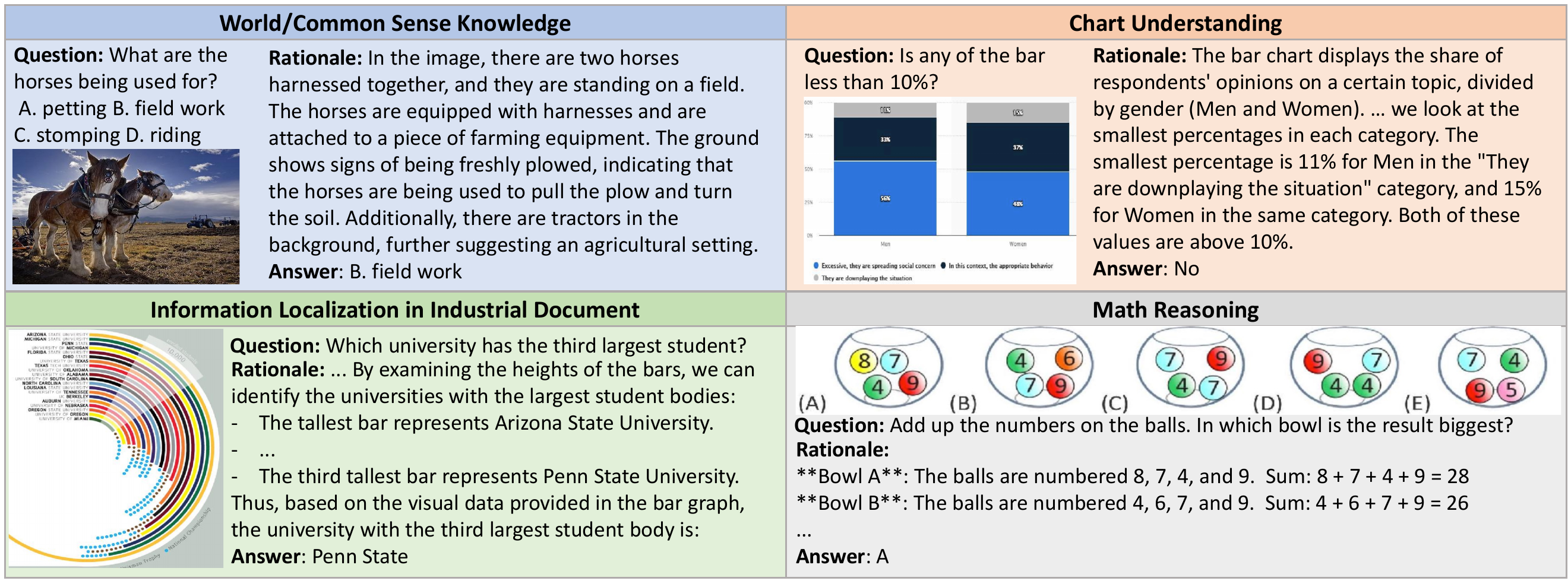}
    \caption{Distillation of examples from various VLM task domains, highlighting the specific reasoning capabilities required.}
    \label{fig:distillation_example}
    \vspace{-10pt}
\end{figure*}

To mitigate the limited availability of high-quality CoT data, we leverage VQA datasets with short annotations and augment them with rationales generated by the \gpto model. We collect 193k visual CoT instances to create the \sharegpto dataset, which we plan to release for public use.  We focus on the following reasoning types as demonstrated in \cref{fig:distillation_example}:

\begin{wrapfigure}{r}{0.4\linewidth}
    \centering
    \includegraphics[width=\linewidth]{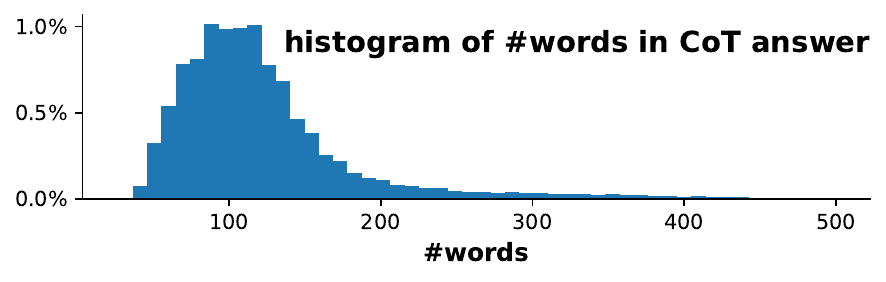}
    \includegraphics[width=\linewidth]{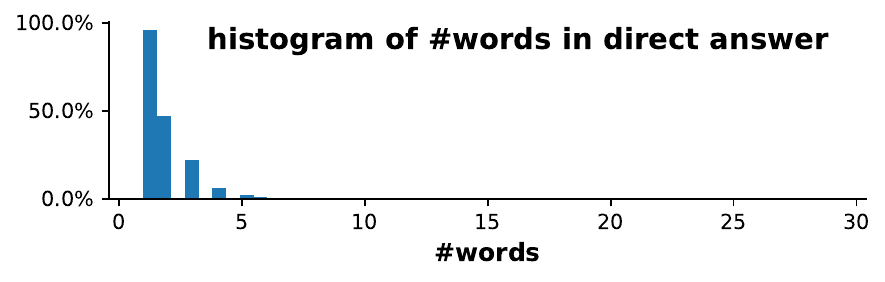}
    \caption{The distribution of word counts for CoT and direct  answer. \label{fig:data_hist}}
    \vspace{-20pt}
\end{wrapfigure}

\noindent \textbf{Real-World Knowledge} includes the \aokvqa dataset~\citep{aokvqa}, which covers a broad range of commonsense reasoning and real-world knowledge for answering questions.

\noindent \textbf{Chart Understanding} includes the \chartqa dataset~\citep{zhang2024tinychart}, which involves tasks like item comparison, counting, and numerical computation.

\noindent \textbf{Document Understanding/Real-World Text} includes \docvqa~\citep{docvqa}, \infovqa~\citep{infographicvqa}, and \textvqa~\citep{textvqa}, focusing on information localization and extraction in industrial documents and real-world image comprehension.

\noindent \textbf{Math and Science} includes \mathvision~\citep{mathvision}, \gllava~\citep{gllava}, \sqa~\citep{scienceqa}, and \aitd~\citep{ai2d}, focusing on scientific knowledge and mathematical reasoning.

After distillation, we filtered out examples whose answer predicted by \gpto is different from ground truth. The data statistics are presented in \cref{tab:cot_data_stats}, and a comparison of answer lengths is shown in \cref{fig:data_hist}, highlighting that CoT responses peak around 100 tokens, while direct answers are typically under 5 tokens. The exact distillation prompt is provided in \cref{appendix:gpto_distill}.

\subsection{Supervised Fine-Tuning for Chain-of-Thought Prediction}
\label{method:sft}
We choose LLaMA3-LLaVA-NeXT-8B as our base architecture, whose weight is initialized with the Open-LLaVA-NeXT weights\footnote{https://github.com/xiaoachen98/Open-LLaVA-NeXT}. To ensure the model handles both direct and chain-of-thought (CoT) predictions, we implement two types of prompts during training.

\textbf{Direct Prediction:} For direct prediction tasks, we use the prompt “Answer the question with a short answer” for short-answer questions, and “Answer with the option’s letter from the given choices directly” for multiple-choice questions.

\textbf{CoT Prediction:} For CoT prediction tasks, we use the prompt “Generate a reason first and then output a letter answer” for multiple-choice questions, and “Generate a reason first and then output a short answer” for short-answer questions. In the model’s response, the rationale is followed by the answer, which is formatted as “\#\#\# Answer: ” to enable answer extraction during evaluation.


\subsection{Reinforcement Learning for Enhanced Reasoning}
\label{method:rl}
To further improve the quality of reasoning chains, we apply RL using the DPO algorithm to better align the model’s reasoning process toward more accurate predictions. The DPO algorithm requires both positive and negative responses. To generate these, we use the SFT model as the policy model (i.e., generator), producing 32 candidate predictions per question (temperature 1.0 for short answer and 1.2 for multiple-choice questions). Each prediction is compared with the ground truth to determine its correctness (\cref{fig:illustration}). Following the approach in \cite{llama3}, we select instances with an accuracy between 0.25 and 0.85. From these, we randomly pair positive and negative responses, creating up to three pairs per question.

Formally, the dataset is denoted as $\mathcal{D}_{DPO} = \{ (\mathcal{V}, x, y_w, y_l )\}$, where $\mathcal{V}$ is the image, $x$ is the question, $y_w$ and $y_l$ are the positive and negative responses. The DPO objective is defined as below:
{\small
\begin{equation*}
 \mathcal{L}_{\mathrm{DPO}}\left(\pi_\theta ; \pi_{\mathrm{ref}}\right) = -\mathbb{E}_{\left(\mathcal{V}, x, y_w, y_l\right) \sim \mathcal{D}_{DPO}} \Bigg[ 
 \log \sigma\Bigg(\beta \log \frac{\pi_\theta\left(y_w \mid x,\mathcal{V} \right)}{\pi_{\text {ref }}\left(y_w \mid x,\mathcal{V}\right)} -\beta \log \frac{\pi_\theta\left(y_l \mid x,\mathcal{V}\right)}{\pi_{\text {ref }}\left(y_l \mid x,\mathcal{V}\right)}\Bigg)\Bigg]\,,
\end{equation*}
}
where $\pi_\theta$ is the policy model to be optimized and $\pi_{\text {ref }}$ is the base reference model, both models are initialized with SFT weights. $\sigmoid$ is the logistic function and $\beta$ is set to $0.1$.



\begin{table}[ht!]
\centering
\caption{
SFT experiments with data composition in \cref{fig:sft_data}: \cOne format alignment only, \cTwo direct responses only, \cThree CoT responses only and \cFour both direct and CoT responses. Inference is performed using both direct and CoT templates. The best CoT prediction result is highlighted in \highc{orange}, while the best direct prediction result is marked in \highd{blue}. The results demonstrate that combining CoT and direct responses during training leads to the best performance across both types of prompts. Refer to \cref{sec:sft} for detailed analysis.
}
\resizebox{1.0\linewidth}{!}{
    \begin{tabular}{lcccccccccc}
    \toprule
    \textbf{Methods}
    & \textbf{Prompting}
    & \aok
    & \chartqa
    & \docvqa 
    & \infovqa
    & \textvqa
    & \aitd
    & \sqa
    & \mathvista
    & Avg
    \\
    \midrule

LLaVA-Next & direct & 85.8 & 70.2 & 75.7 & 37.7 & 68.2 & 71.5 & 75.4 & 39.3 & 65.5\\ 
+ Format \cOne & CoT & 84.3 & 71.2 & 67 & 34.9 & 62.2 & 67.4 & 74.4 & 40.3 & 62.7 \\ 

\hline 

LLaVA-Next & direct & \highd{86.4}	& 73.7	& 78 & 45.4 & 71.9 & 78.8 & \highd{91.5} & 43.2 & 71.1 \\
+ Direct \cTwo & CoT & 85.7 & 71.8 & 68.8 & 38.6 & 63.6 & 72.5 & 85.4 & 38.6 & 65.6 \\
    
\hline

LLaVA-Next & direct & 84.9 & 71.8 & 81.2 & 45.7 & 72.1 & 75.3 & 85 & 41.9 
 & 69.7 \\
+ Cot \cThree & CoT &  85.1 & 82.2 & 81.2 & 49.7 & 69.9 & 77 & 91.3 & 49.2 &  73.2 \\

\hline

LLaVA-Reasoner  & direct & 85.4 & \highd{76.1} & \highd{82.9} & \highd{50.6} & \highd{73.1} & \highd{79.4} & 90.4 & \highd{44.3} & \highd{72.8} \\
-SFT \cFour & CoT & \highc{86.2}	& \highc{83.0}	& \highc{81.8}	 & \highc{51.6}	 & \highc{71.1}	 & \highc{78.5}	 & \highc{92.7} & \highc{50.6} & \highc{74.4} \\

\bottomrule
\hline
\end{tabular}
}

\label{tab:sft_result}
\end{table}

\begin{wrapfigure}{r}{0.5\linewidth}
    \centering
    \vspace{-20pt}
    \includegraphics[width=\linewidth]{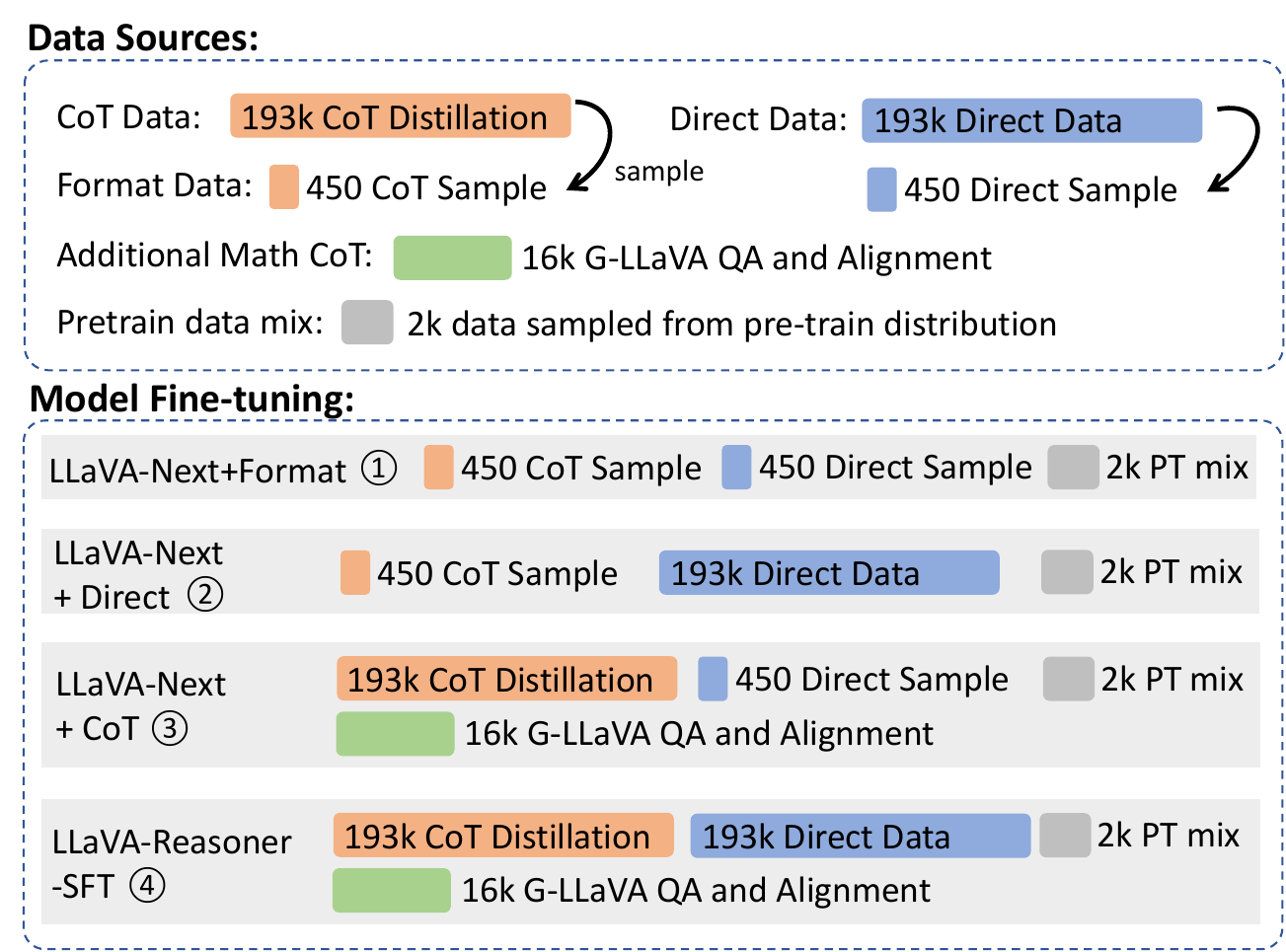}
    \caption{The upper section displays the data sources used for the SFT experiments, while the lower section illustrates the data composition for model training.    \label{fig:sft_data}}
    \vspace{-30pt}
\end{wrapfigure}



\section{SFT Experiments for Chain-of-thought Learning}
\label{sec:sft}
In this section, we explore how SFT can enhance VLM reasoning by addressing two key research questions: (1) \textit{Can CoT reasoning be implicitly learned from short responses?} and (2) \textit{How effectively can CoT be learned from \gpto distilled data?} Additionally, we analyze the composition of CoT data across various reasoning capabilities and compare the performance of SOTA models with \gpto.

\subsection{Training Setting}
As shown in the upper part of \cref{fig:sft_data}, we present the data composition for SFT. The training data includes CoT distillation (193k instances) from \cref{tab:cot_data_stats} and corresponding short answers (193k). Additionally, for CoT data, we incorporate 16k visual math examples from \gllava. To maintain general instruction-following capability as the base model, we include 2k randomly sampled instruction data from LLaVA pretraining ~\cite{liu2024llavanext}. To ensure the SFT models can handle both direct and CoT prompts during inference, we sample a small set of format-aligned data—50 examples from each of the 9 datasets—resulting in 450 instances.

In the lower part of \cref{fig:sft_data}, we outline the data composition for model training. Specifically, \modelformat (\cref{fig:sft_data} \cOne) serves as the baseline model, trained exclusively on format-aligned data to enforce the desired output format without learning any task-specific reasoning skills. In contrast, models in \cref{fig:sft_data} \cTwo and \cThree incorporate either direct or CoT datasets, enabling the model to be expert in one type of skill as well as following the both direct and CoT prompt styles. Finally, \modelsft (\cref{fig:sft_data} \cFour) represents the SFT model trained on both CoT and direct data, making it to be expert in both types of reasoning.

We use the LLaMA3-LLaVA-NeXT-8B architecture, initializing the weights with Open-LLaVA-NeXT. All Supervised Fine-Tuning (SFT) experiments are trained for 1 epoch with a learning rate of 5e-6 and a batch size of 32. The experiments are conducted on 8 H100 GPUs.

\subsection{Evaluation Setting}
We evaluate our method using a range of benchmark datasets, including \aokvqa~\citep{aokvqa}, \chartqa~\citep{chartqa}, \docvqa~\citep{docvqa}, \infovqa~\cite{infographicvqa}, \textvqa~\citep{docvqa}, \aitd~\citep{ai2d}, \scienceqa~\citep{scienceqa}, and \mathvista~\citep{mathvista}. We also conduct more evaluation on general datasets \ocrbench~\citep{ocrbench}, \mmstar~\citep{mmstar}, and \mmmu~\citep{mmmu} in later sections.
The evaluation for \aokvqa was implemented by us, while for the other datasets, we follow the evaluation protocols outlined in VLMEval~\citep{duan2024vlmevalkit}. 

For CoT evaluation, answers are extracted after the pattern "\#\#\#Answer: " before sent to evaluation. More comparison with LLaMA3-LLaVA-NeXT-8B model is shown \cref{appendix:baseline_eval} and evaluation on GPT-4o is shown in \cref{appendix:gpto_eval}.

\subsection{Can reasoning be implicitly learnt from direct prediction?}
\Cref{tab:sft_result} presents the performance of the models introduced in \cref{fig:sft_data}. Since \modelbase training data contains very few CoT reasoning examples, CoT performance of \cOne lags behind direct prediction across most tasks. The only improvement is observed in \chartqa and \mathvista with a modest gain of +1.0  in CoT performance, showing CoT is helpful for calculation related tasks.

When comparing model trained on direct only data (\cTwo) to that trained on format-aligned data (\cOne), we observe an average gain of +5.6 in direct prediction accuracy (65.5 $\rightarrow$ 71.1) and a +2.9 improvement in CoT performance (62.7 $\rightarrow$ 65.6). Surprisingly, closer inspection of CoT performance in calculation-involved tasks, such as \chartqa and \mathvista, reveals only marginal gains (+0.6 for \chartqa CoT) or even a performance drop (-1.7 on \mathvista), which contrasts with the improvements seen on the two tasks in \cOne. 
On text-rich tasks, positive gains (>1) are observed, with the most improvement seen in \infovqa (+3.7). Significant gains are also evident in science-related tasks like \aitd (+5.1) and \sqa (+11.0). Despite these improvements, CoT performance still trails behind direct prediction overall (CoT: 65.6 vs. direct: 71.1). This result suggests that training on direct only prediction may not effectively help with CoT prediction.

\begin{wraptable}{r}{0.4\linewidth}
\small
\vspace{-15pt}
\caption{Effect of data composition on math reasoning. MV: \mathvision, GL: \gllava, MI: MathInstruct, MP: MathPlus. 
\label{tab:ablation_math}
}
\begin{tabularx}{\linewidth}{lc}
\hline
\textbf{Data Config} & \makecell{\bf\mathvista \\ (direct/CoT)} \\ \hline
format only \cOne & 39.3/40.3 \\
MV & 41.0/43.4 \\
MV+GL &  43.2/44.9 \\
MV+GL+MP50k & 42.3/45.6 \\
MV+GL+MP100k & 43.0/44.9 \\
MV+GL+MI50k & 43.1/45.0 \\
MV+GL+MI100k & 43.7/46.3 \\
MV+GL+\aitd & 44.1/46.4 \\
MV+GL+\sqa & 43.1/47.3 \\
MV+GL+\chartqa & 43.2/50.4
\end{tabularx}
\vspace{-10pt}
\end{wraptable}

\subsection{How Effective is CoT Reasoning Data?}
When comparing the model trained on CoT-only data (\cThree) with the one trained on format-aligned data (\cOne), we observe improvements in both direct and CoT predictions. Direct prediction performance increases by an average of +4.2 (65.5 $\rightarrow$ 69.7), while CoT prediction improves significantly by +10.5 (62.7 $\rightarrow$ 73.2). Notably, the CoT performance of the model \cThree surpasses its direct prediction (73.2 CoT vs. 69.7 direct). Significant gains are observed in calculation-intensive tasks like \chartqa and \mathvista, with increases of +11.0 and +8.9 in CoT performance, respectively. Interestingly, for text-rich tasks such as \docvqa, \infovqa, and \textvqa, the direct performance of model \cThree (trained on CoT-only data) outperforms that of model \cTwo (trained on direct-only data). This suggests that even for text-heavy tasks, reasoning processes, such as localizing information in documents or recognizing text in real-world scenarios, may benefit from CoT training. The skills learned from CoT training appear to generalize to direct prediction as well.

\begin{wraptable}{r}{0.4\linewidth}
\small
\vspace{-10pt}
\caption{Effect of data composition on science related tasks.
\label{tab:ablation_sci}
}
\begin{tabularx}{\linewidth}{lcc}
\hline
\textbf{Data Config} & \bf\aitd & \bf \sqa \\ \hline
format only \cOne & 67.4 & 74.4 \\
\aitd & 76.3 & 76.6 \\
\sqa & 66.9 & 90.4 \\
\aitd+\sqa &  76.7 & 91.2 \\
\aitd+\sqa+\chartqa & 77.4 & 91.4 \\
\end{tabularx}
\vspace{-10pt}
\end{wraptable}

When both CoT and direct data are combined (\cFour), performance is further enhanced for both prediction types, with an average gain of +7.3 in direct prediction (65.5 $\rightarrow$ 72.8) and +11.7 in CoT prediction (62.7 $\rightarrow$ 74.4). This demonstrates that combining direct and CoT data yields the best overall performance. Interestingly, in model \cFour, for 3 out of 8 datasets (\textvqa, \docvqa, \aitd), direct prediction outperforms CoT prediction. We hypothesize that these tasks involve a significant proportion of concise fact extraction, where generating long-form CoT responses may not provide additional benefits or even hurts. Further validation of this hypothesis will be explored in future work.

\subsection{Ablation Tests on Data Composition}
\paragraph{Data Composition for Math.} In \cref{tab:ablation_math}, we examine the effectiveness of data composition on \mathvista performance. We first include two visual math datasets: \mathvision (MV) and \gllava (GL). Including MV improves CoT performance by +3.1 over format only baseline (\cref{fig:sft_data} \cOne), while adding GL yields an additional gain of +1.5. Building on MV+GL, we incorporate several datasets that are potentially relevant to the task, including two math text-only datasets: MathPlus (MP) and MathInstruct (MI), two science datasets: \sqa and \aitd, and \chartqa. Notably, \chartqa significantly boosts CoT performance (+5.5), while \aitd and \sqa provide positive gains of +0.6 and +1.5, respectively. However, adding the math text datasets results in minimal improvement. Comparing inclusion of 100k MP vs 50k MP, more text data does not necessarily lead to better results. Therefore, we decided not to include them in training \modelsft.

\paragraph{Data Composition for Science Tasks with CoT Prediction.} In \cref{tab:ablation_sci}, we evaluate the impact of data composition on science datasets, including \aitd and \sqa. Our results show that combining \sqa and \aitd provides additional gains on both datasets, indicating that they are mutually beneficial. Furthermore, adding \chartqa contributes positively to both datasets, with a notable improvement of +0.7 for \aitd.

\begin{wraptable}{r}{0.52\linewidth}
\vspace{-10pt}
\caption{
Performance Comparison of \gpto, Cambrian-7b, and our SFT Model. For Cambrian, * indicates our replicated results, while others are adapted from \cite{tong2024cambrian}, $^\dagger$ indicate CoT prompt used for evaluation. ‘Our-SFT’ refers to \modelsft.
}
\centering
\resizebox{1.0\linewidth}{!}{
    \begin{tabularx}{\linewidth}{l|c|cc}
    \toprule
    \textbf{Dataset} & \gpto & Cambrian & Our-SFT  \\
    & direct/cot & official & direct/cot\\
    \midrule
    \aok & 89.6/90.1 & 83.1* & 85.4/86.2  \\
    \chartqa & 79.6/84.7 & 73.3 & 76.1/83.0 \\
    \docvqa & 90.3/90.8 & 77.8 & 82.9/81.8   \\
    \infovqa & 72.4/72.8 & 45.7* & 50.6/51.6  \\
    \textvqa & 78.1/75.4 & 71.7 & 73.1/71.1   \\
    \aitd & 80.7/81.5 & 73.0 & 79.4/78.5  \\
    \sqa & 85.9/87.2 & 80.4 & 90.4/92.7   \\
    \mathvista & 54.8/63.4 & 49.0$^\dagger$ & 44.3/50.6  \\
    \ocrbench & 80.2/79.2 & 62.4 &  61.6/62.0  \\
    \mmstar & 55.1/64.7 & 50.3* & 51.6/54.0  \\
    \mmmu & 57.8/63.6 & 42.7 & 41.6/40.0  \\
    \hline
    Avg (of best) & 77.9 & 64.5 & 68.8 \\
    \bottomrule
    \end{tabularx}
}
\vspace{-10pt}
\label{tab:sota_compare}
\end{wraptable}

\subsection{Comparing with SOTA model and \gpto}
In \cref{tab:sota_compare}, we compare the performance of \gpto and a recent state-of-the-art model, Cambrian \cite{tong2024cambrian}. For \gpto, we include both direct and CoT predictions, following the prompt optimization steps outlined in \cite{borchmann2024notes}, with the prompts detailed in \cref{appendix:gpto_eval}. For Cambrian, we report the numbers from \cite{tong2024cambrian} and replicated the results using the official checkpoint on \mmstar, \infovqa, and \aokvqa. Specifically for Cambrian, CoT predictions were used for the \mathvista dataset, while direct predictions were applied for the remaining datasets. 

When compared to open-source models, \gpto outperforms on nearly all benchmark datasets, with the exception of \sqa. Notably, significant improvements from CoT predictions are observed on tasks involving calculation or complex reasoning, such as \chartqa, \mathvista, \mmmu, and \mmstar.

Cambrian-7B is trained on a dataset of 7 million open-source instruction-following examples. In contrast, our model, fine-tuned on fewer than 400k instruction examples, outperforms Cambrian-7B on most benchmark datasets, underscoring the effectiveness of incorporating CoT data. While we recognize the challenge of comparing against other models, such as One-Vision~\citep{onevision}, MiniCPM-V~\cite{minicpm}, X-Composer~\cite{xcomposer}, and InternVL~\cite{internvl}, due to differences in model architecture, training datasets, and evaluation pipelines, our primary focus is on studying the effectiveness of CoT learning rather than competing for state-of-the-art performance on visual-language tasks.

\begin{table}[ht!]
\centering
\caption{
DPO experiment with \modelsft as the base policy model. We compare two DPO datasets: \cFive RLAIF-V~\cite{yu2024rlaif} and \cSix our preference dataset comprising \aokvqa, \chartqa, and math. The best CoT prediction is highlighted in \highc{orange}. Our DPO dataset shows the better improvements in chain-of-thought reasoning.
}
\resizebox{1.0\linewidth}{!}{
    \begin{tabular}{lcccccccccc}
    \toprule
    \textbf{Methods}
    & \textbf{Prompting}
    & \aok
    & \chartqa
    & \docvqa 
    & \infovqa
    & \textvqa
    & \aitd
    & \sqa
    & \mathvista
    & Avg
    \\
    \midrule

LLaVA-Reasoner  & direct & 85.4 & 76.1 & 82.9 & 50.6 & 73.1 & 79.4 & 90.4 & 44.3 & 72.8 \\
-SFT \cFour & CoT & 86.2	& 83.0	& 81.8	& 51.6	 & 71.1	 & 78.5	 & 92.7 & 50.6 & 74.4 \\

\hline

LLaVA-Reasoner  & direct & 85.6	& 76.1 & 83.1 & 50.7 & 73.3	 & 79.6 & 91.1 & 44.1 & 73.0 \\
-RLAIF \cFive & CoT & 86.7 & 83.0 & 82.4 & 50.8 & 71.4 & 79.1 & \highc{92.9} & 50.8 & 74.6 \\

\hline 

LLaVA-Reasoner  & direct & 85.4	& 76.4 & 83.1 & 51.2 & 73.3 & 79.4 & 90.8 & 44.2 & 73.0 \\
-DPO-ours \cSix & CoT & \highc{87.0} & \highc{84.2} & \highc{82.7} & \highc{52.7} & \highc{71.5} & \highc{79.5} & 92.6 & \highc{52.1} & \highc{75.3} \\
\bottomrule
\hline
\end{tabular}
}

\label{tab:dpo_result}
\end{table}

\section{RL Experiments for Enhanced Chain-of-thought Reasoning}

In this section, we demonstrate the effectiveness of RL in further enhancing CoT reasoning. We employ the DPO algorithm, which is directly optimized using positive and negative pairs. By leveraging short-answer feedback (\cref{method:rl}), we construct preference pairs across three domains: \aokvqa (real-world knowledge reasoning), \chartqa (chart interpretation), and math (\mathvision and \gllava). Although additional DPO data from other datasets could be incorporated, data scaling and balancing will be addressed in future work.

For the DPO dataset, we include 24.5k examples from \chartqa, 18.3k from \aokvqa, and 22.0k from math domain, totaling 64.8k preference data pairs. We train \modelsft on this dataset using a learning rate of 5e-7, a batch size of 32, and for 1 epoch. We found an additional trick to truncate the responses up to 90 tokens to be helpful for DPO training. To compare the effectiveness of different DPO datasets, we include RLAIF-V~\cite{yu2024rlaif}, which contains 80k DPO pairs representing the state-of-the-art dataset for aligning VLMs for reducing hallucinations.

\subsection{Can DPO Calibrate Reasoning?}
In \cref{tab:dpo_result}, we present the results of the DPO model optimized on top of \modelsft (\cFour). Model \cFive uses the SOTA RLAIF-V~\cite{yu2024rlaif} data, while model \cSix uses our dataset. We observe that Model \cFive shows a slight improvement in both direct prediction (+0.2) and CoT prediction (+0.2), whereas model \cSix demonstrates a greater improvement in CoT prediction (+1.1) with equal gains on direct prediction. Interestingly, though only 3 out of 8 datasets are selected to construct DPO pairs, gains are observed across 7 out of 8 datasets except for \sqa with a slight decrease (92.9 $\rightarrow$ 92.6). These results suggest that DPO dataset constructed from model-generated rationales can effectively enhance reasoning accuracy and show generalization across tasks.

\begin{figure*}[ht]
    \centering
    \includegraphics[width=0.32\linewidth]{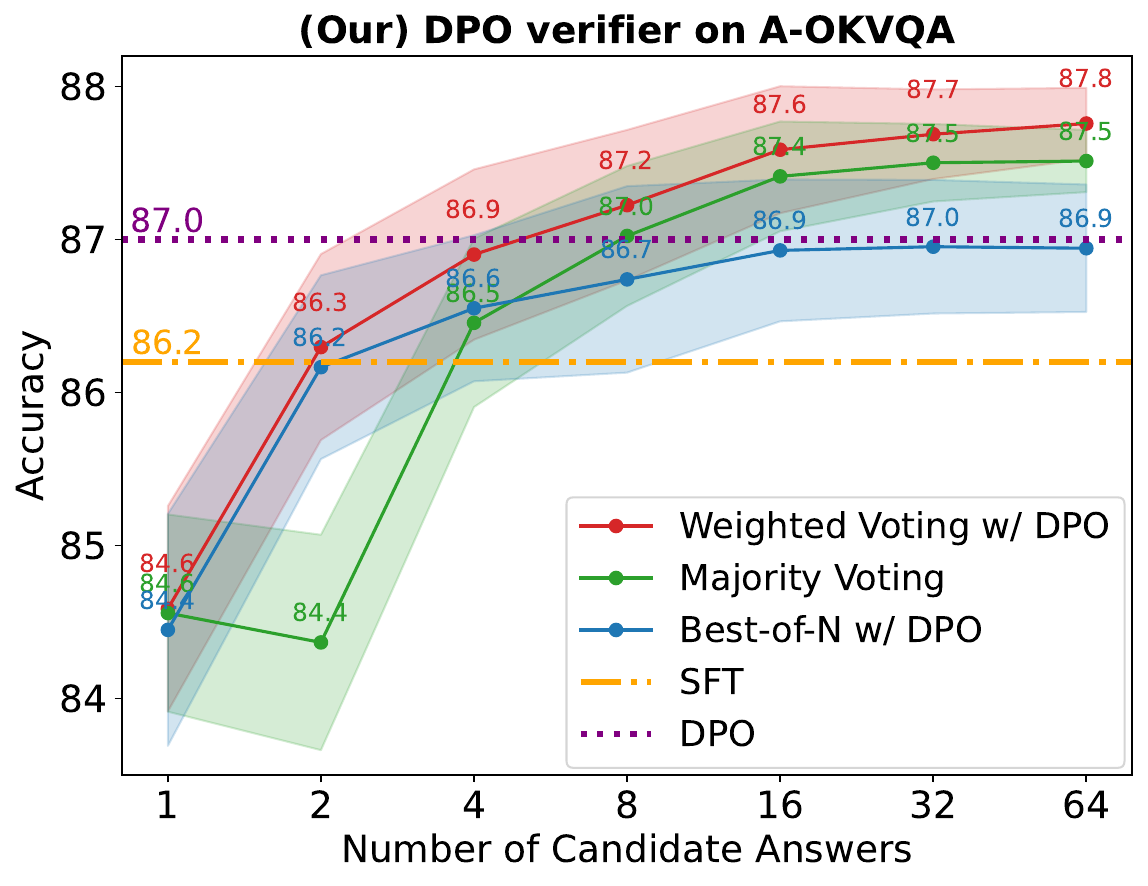}
    \includegraphics[width=0.32\linewidth]{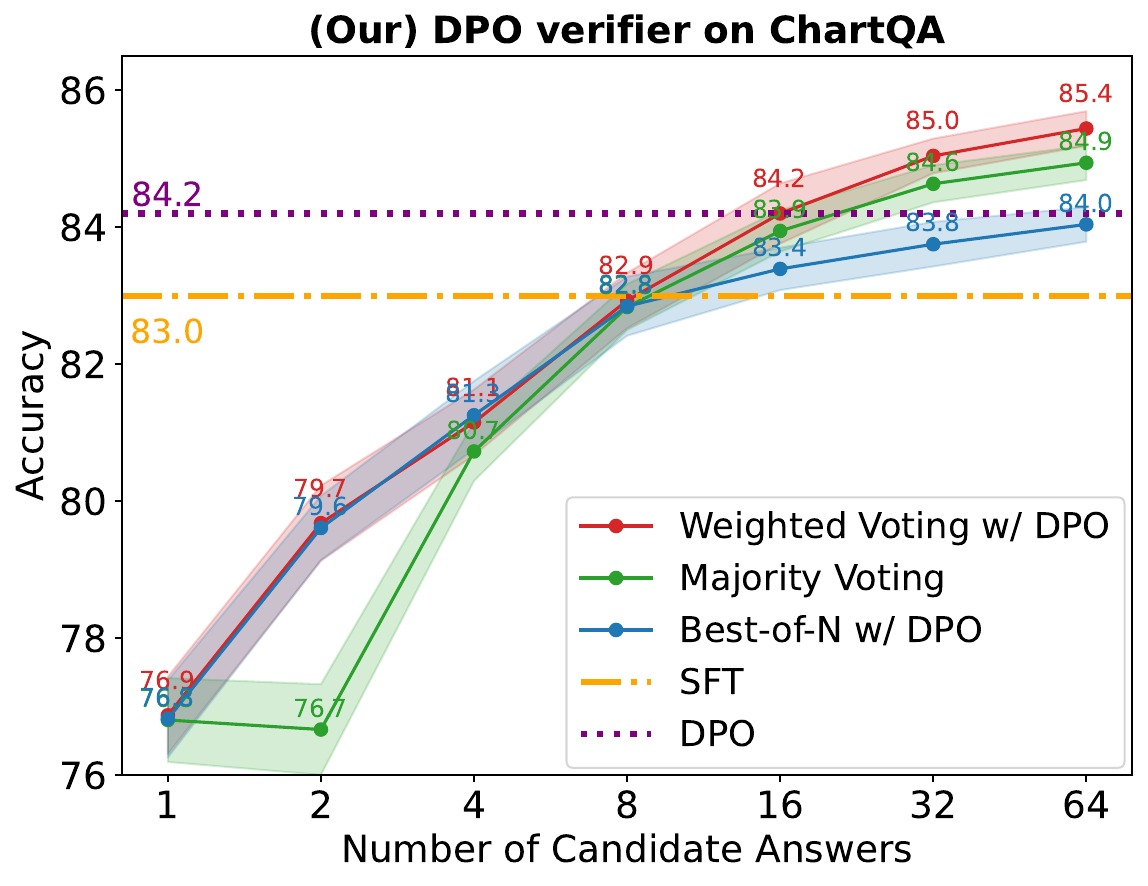}
    \includegraphics[width=0.33\linewidth]{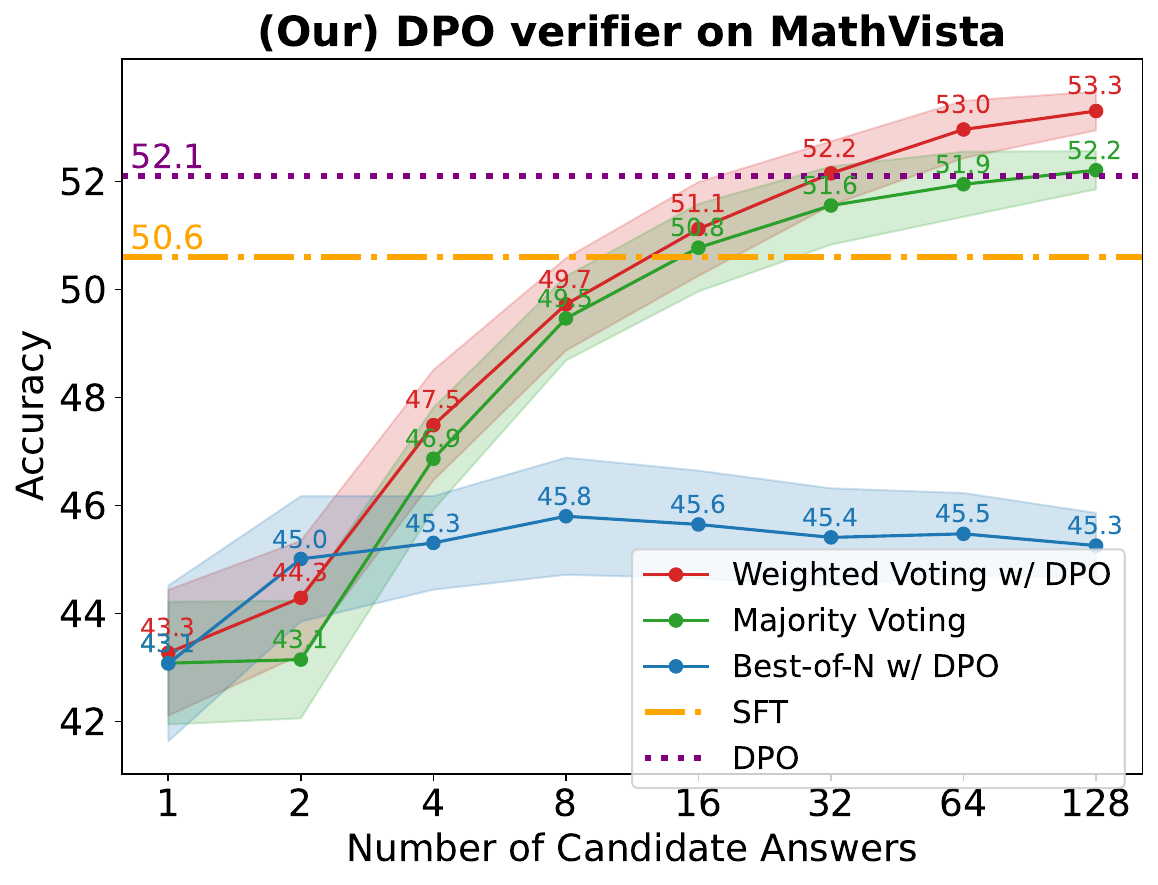}
        \includegraphics[width=0.32\linewidth]{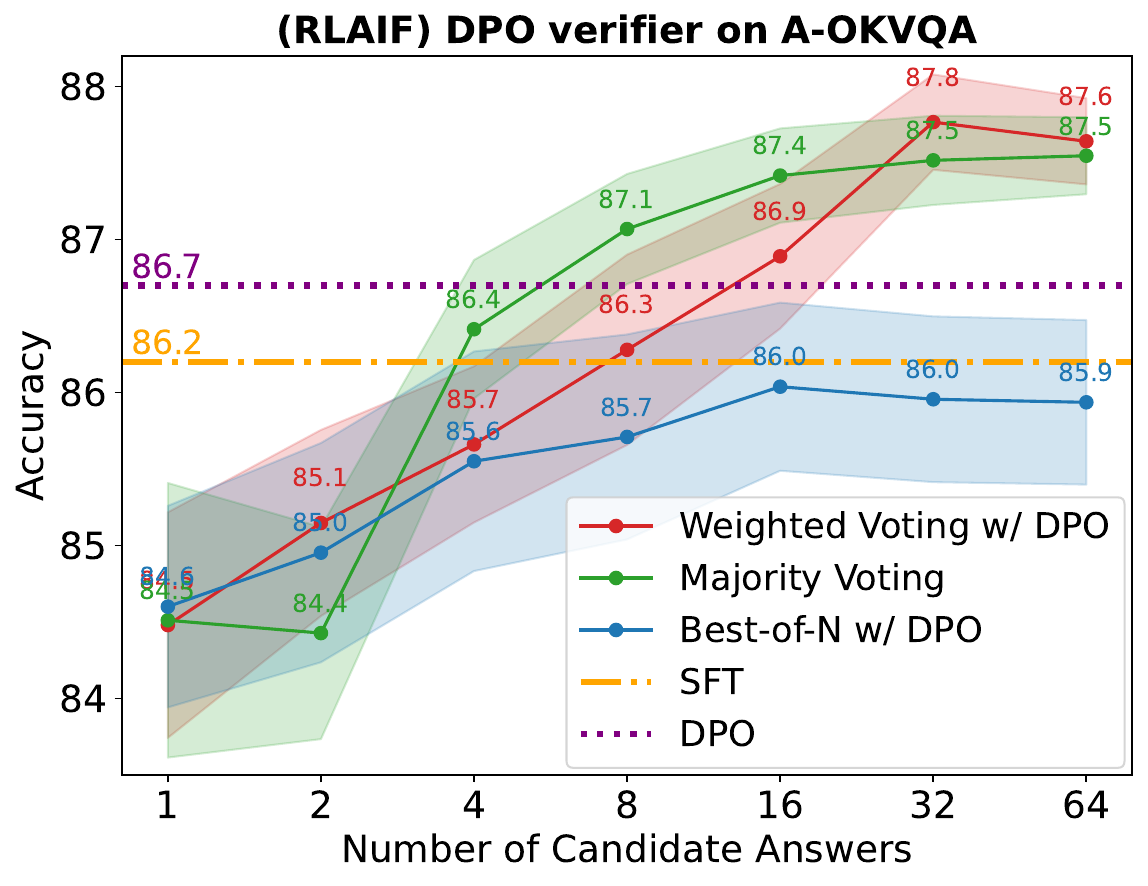}
    \includegraphics[width=0.32\linewidth]{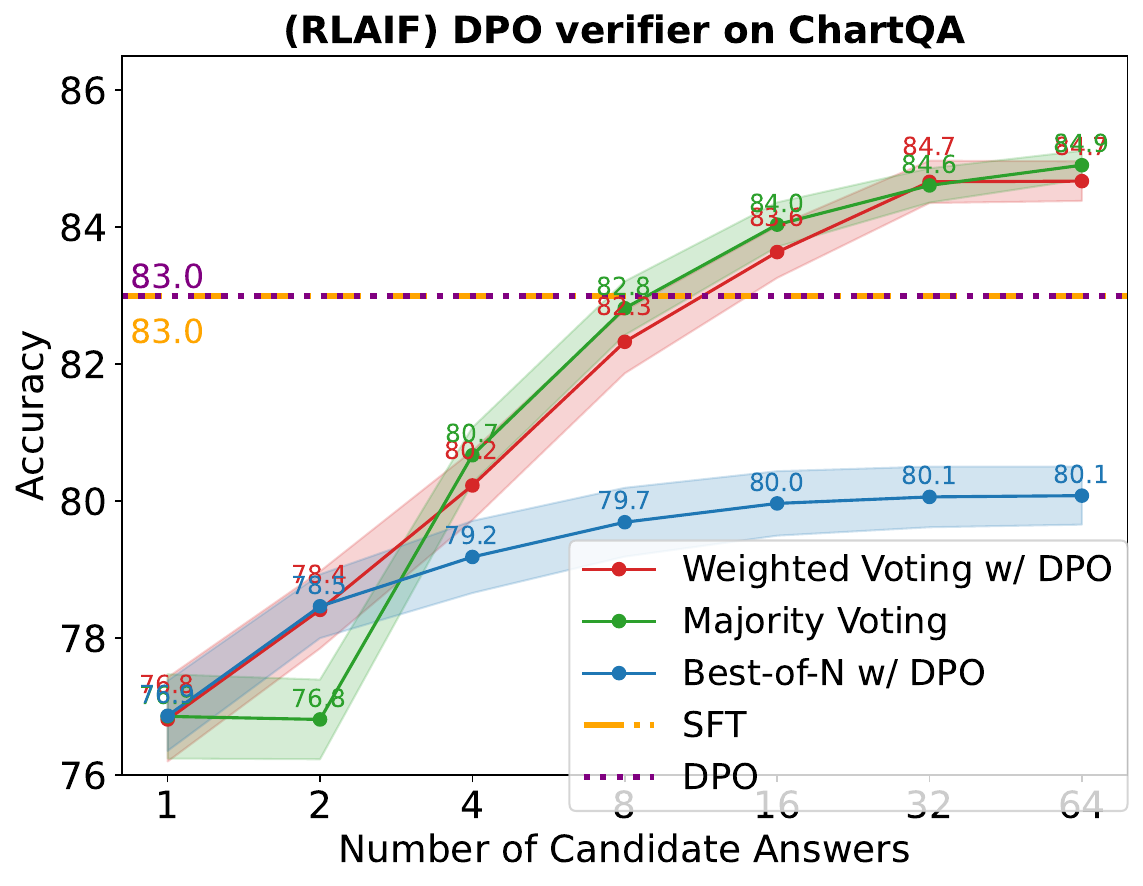}
    \includegraphics[width=0.33\linewidth]{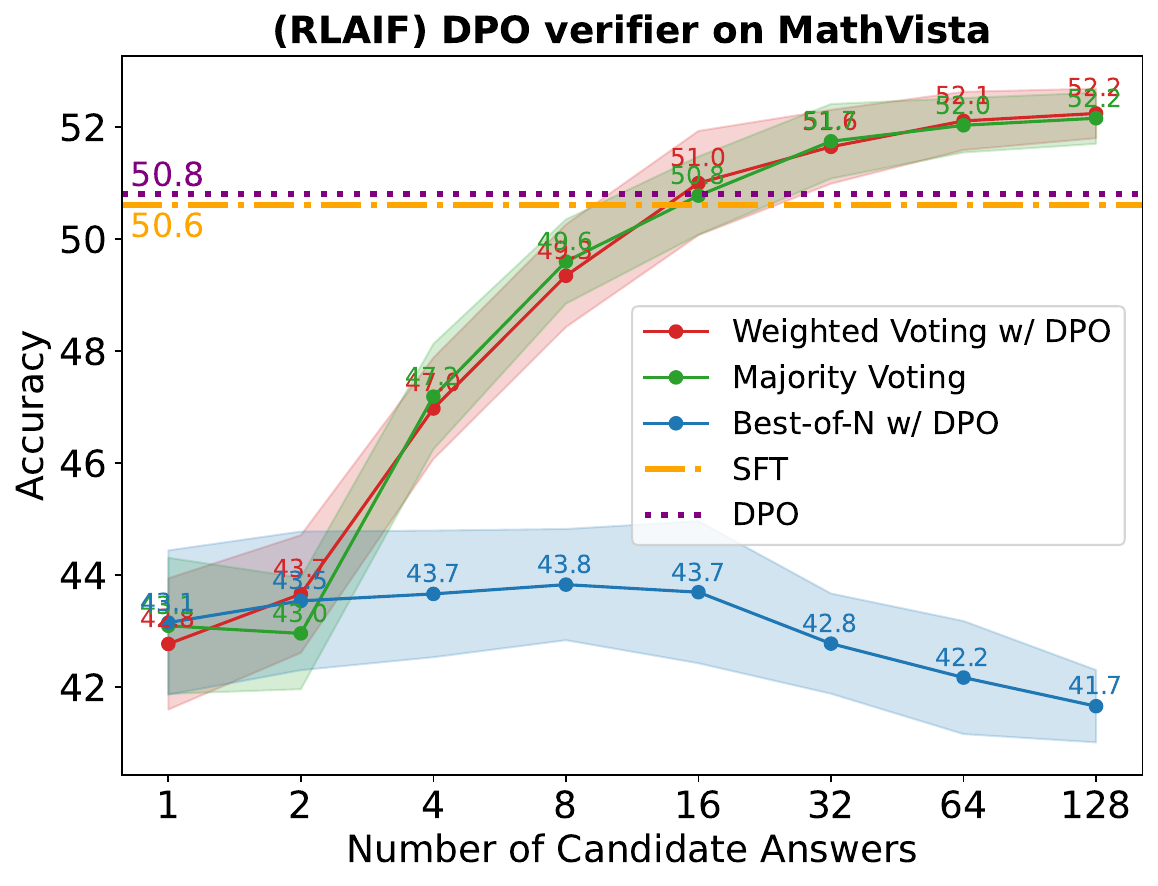}
    \caption{The figures illustrate the performance of the DPO model as a verifier on \chartqa, \aokvqa, and \mathvista. Compared to the model trained with RLAIF-V, the model trained on our reasoning data pairs consistently shows improvement in both best-of-N selection and weighted voting.}
    \label{fig:dpo_verifier}
\end{figure*}


\subsection{DPO as Verifier for CoT Reasoning Re-ranking}
\vspace{-5pt}
In \cref{fig:dpo_verifier}, we present the re-ranking results using the DPO model as a verifier, following the approach of \cite{zhang2024direct, vstar, lu2024step}. The DPO reward score is calculated as $\log \frac{\pi_{\text{dpo}}\left(y \mid x,\mathcal{V} \right)}{\pi_{\text{sft}}\left(y \mid x,\mathcal{V}\right)}$, where $\mathcal{V}$ represents the image, $x$ the question, and $y$ the candidate answer. We explore two re-ranking strategies: Best-of-N and Weighted Voting. A Majority Voting (or self-consistency) baseline is also included for comparison. 

\begin{wraptable}{r}{0.65\linewidth}
\vspace{-10pt}
\caption{
More DPO results on general evaluation benchmark datasets.
\label{tab:dpo_general}
}
\centering
\resizebox{1.0\linewidth}{!}{
    
\begin{tabularx}{\linewidth}{lcccc}
\toprule
\textbf{Methods}
& \ocrbench
& \mmstar
& \mmmu 
& Avg
\\
\midrule
SFT \cFour & 62.0 & 54.0 & 40.1 & 52.0 \\
SFT+RLAIF \cFive & \bf 63.7 & 53.5 & 42.3 & 53.2 \\
SFT+DPO-ours \cSix & \bf 63.7 & \bf 54.1 & \bf 42.6 & \bf 53.5 \\
\bottomrule
\end{tabularx}
}
\end{wraptable}

When trained with RLAIF-V data (\cFive), the DPO model demonstrates improvements as both a generator and verifier on \aokvqa, likely due to the dataset’s alignment with real-world images, which matches the nature of \aokvqa. Interestingly, while model \cFive does not show improvements as a generator on \chartqa, it still produces positive results in best-of-N re-ranking, indicating that the learned preferences can generalize across domains. However, weighted voting does not lead to any improvements, and no significant gains are observed in re-ranking for \mathvision. In contrast, when trained with reasoning data pairs, \modelrl (\cSix) shows improvements across both re-ranking metrics, underscoring the effectiveness of DPO on reasoning data pairs.


\vspace{-5pt}
\subsection{Additional DPO CoT Performance on General Datasets}

\begin{wrapfigure}{r}{0.45\linewidth}
    \centering
    \vspace{-20pt}
    \includegraphics[width=\linewidth]{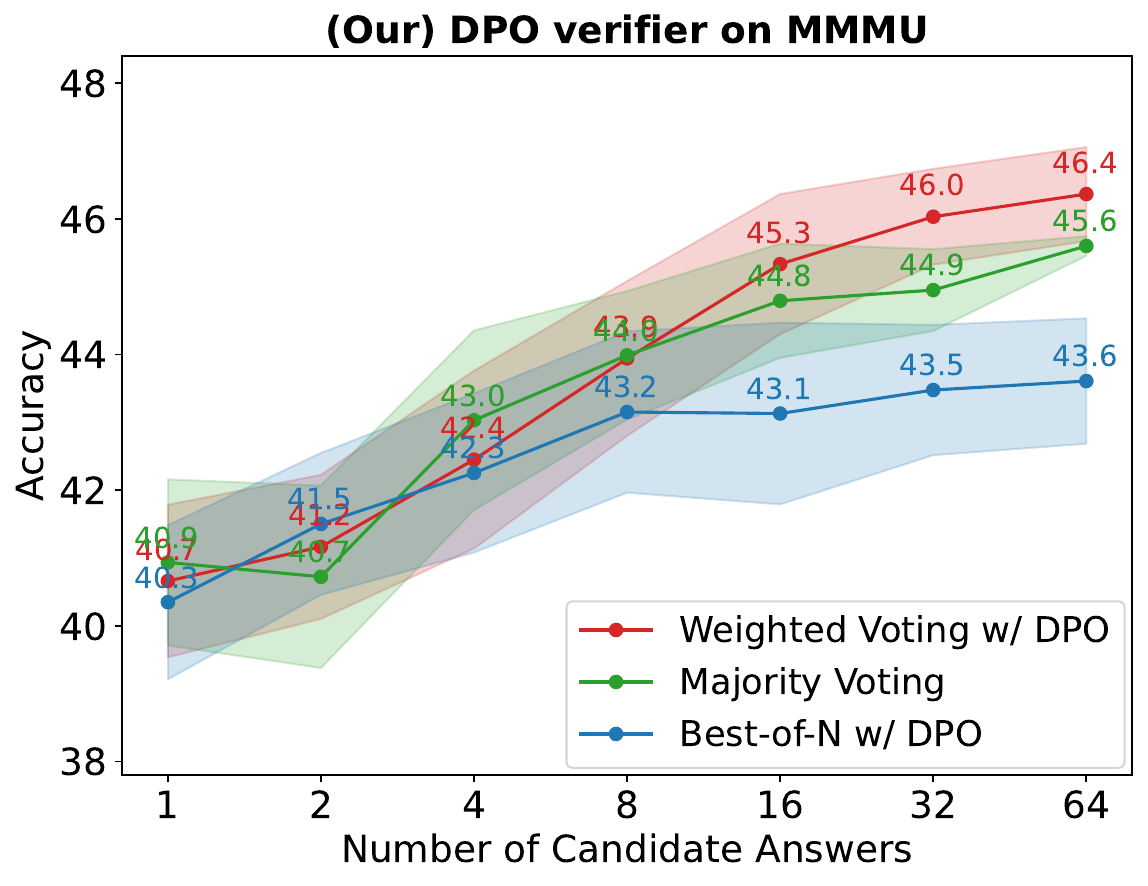}
    \includegraphics[width=\linewidth]{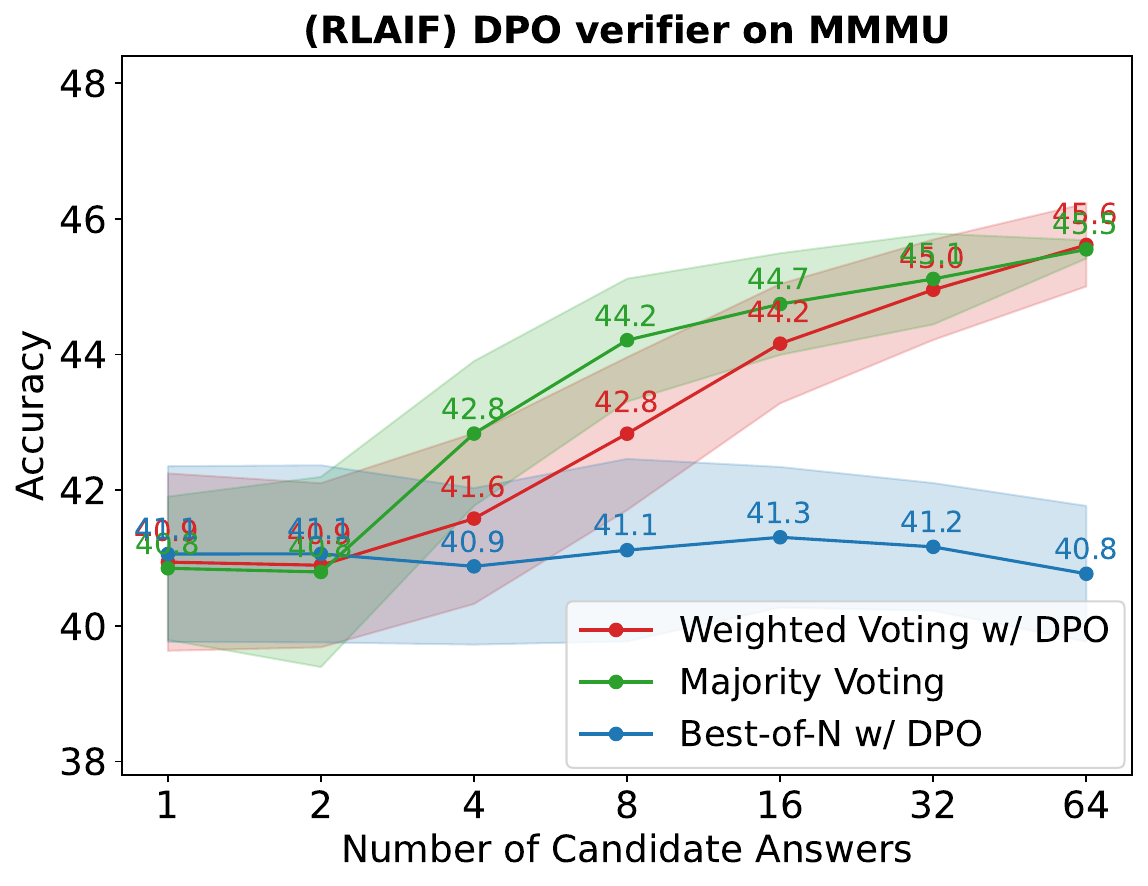}
    \caption{\footnotesize The performance of the DPO verifier on the MMMU dataset with 988 multiple-choice questions. We observe that DPO trained on our reasoning dataset achieves consistent improvements in re-ranking metrics, while DPO trained with RLAIF does not show significant gains.}
    \vspace{-40pt}
    \label{fig:mmmu_verifier}
\end{wrapfigure}

In \cref{tab:dpo_general}, we present the DPO CoT performance on \ocrbench, \mmstar, and \mmmu. We observe that both DPO models outperform the SFT baseline, with our DPO model trained on CoT reasoning pairs showing slightly better results.

In \cref{fig:mmmu_verifier}, we further explore the effectiveness of DPO on the MMMU dataset, which consists of challenging college-level subject questions. We provide re-ranking results for multiple-choice problems from the Dev+Val split (988/1050). First, the SFT model with self-consistency shows consistent improvements reaching 45.5 with 64 candidate votes. \modelrl, trained on reasoning data pairs, shows strong generalization on MMMU by excelling in both weighted voting and best-of-N voting during candidate re-ranking.
While the DPO model trained on RLAIF-V (\cFive) improves CoT predictions, it does not achieve gains in the re-ranking metrics, indicating limitations in distinguishing correct from incorrect reasoning on more complex data. We hypothesize that, compared to \chartqa, the reasoning questions in MMMU are more challenging and span a broader range of subjects. The RLAIF-V dataset relies primarily on COCO image domain, which may not provide sufficient coverage, leading to weaker performance in re-ranking.

\begin{figure*}[ht]
    \centering
    \includegraphics[width=0.95\linewidth]{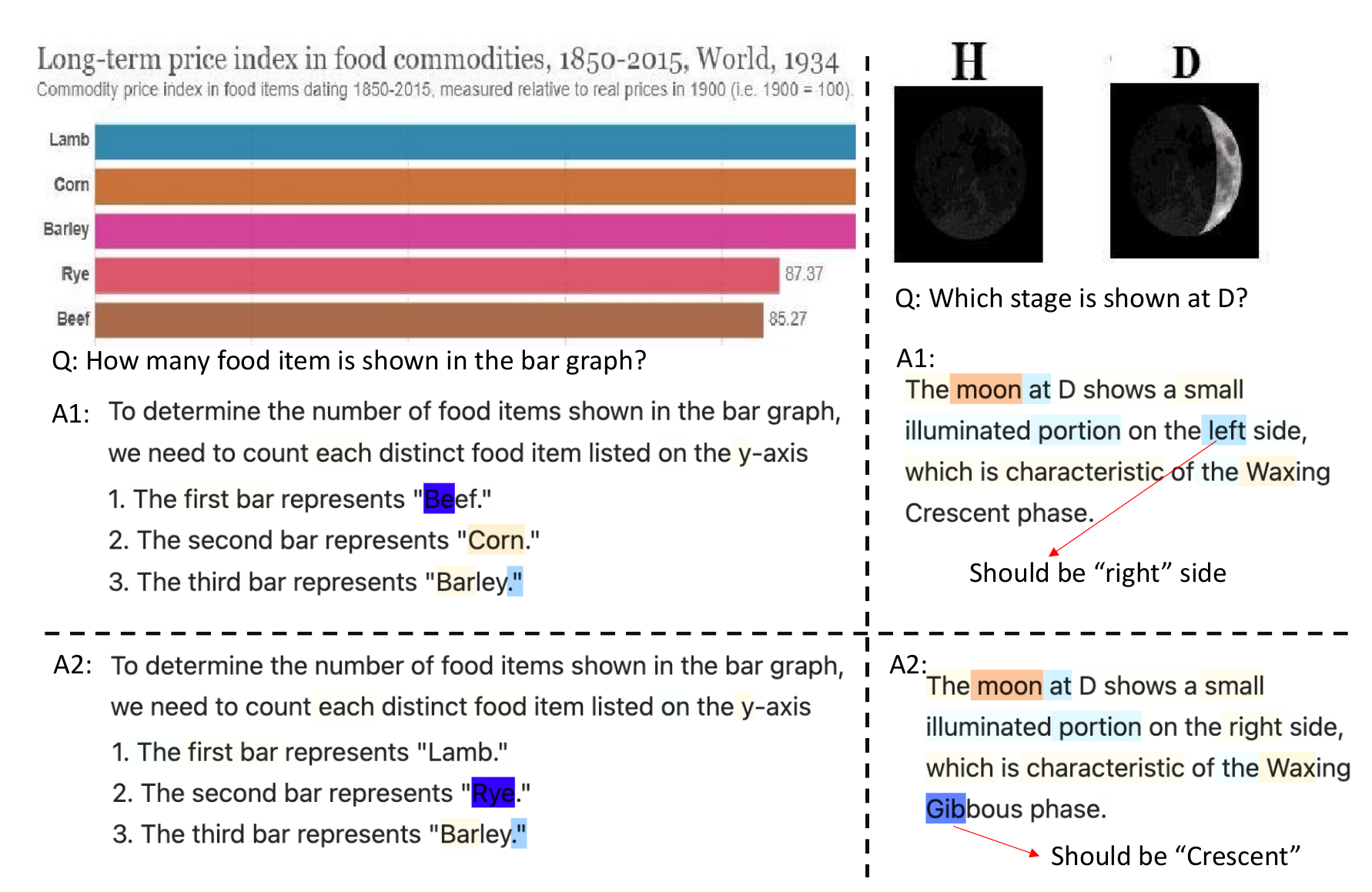}
    \caption{Credit assignment of the DPO model on a portion of the responses from the \chartqa and \aitd datasets. The DPO token-level reward is computed for each token, with the rewards normalized to have a mean of 0. Negative scores are highlighted in cool colors (\highd{blue}), while positive scores are highlighted in warm colors (\highc{orange}). We observe that the DPO model is particularly sensitive to the first mistakes or hallucinations introduced in the response.
}
    \label{fig:dpo_credit_assignment}
\end{figure*}

\subsection{DPO Credit Assignment}

While the DPO model is trained on pairwise data, prior works~\citep{rafailov2024r, lu2024step} have shown that DPO policies can learn to predict \textit{token-level rewards} from binary preference data. These experiments primarily focused on math reasoning with LLMs. In this work, we provide examples of credit assignment learned by the VLM DPO, as shown in \cref{fig:dpo_credit_assignment}. The token-level DPO reward can be expressed as
$\log \frac{\pi_{\text{dpo}}\left(y_i \mid x,\mathcal{V} \right)}{\pi_{\text{sft}}\left(y_i \mid x,\mathcal{V}\right)}$, where $\mathcal{V}$ represents the image, $x$ the question, and $y_i$ the $i$-th token in the generated response. This reward reflects the relative confidence of the DPO model compared to the SFT model for a given token in a candidate response.

In \cref{fig:dpo_credit_assignment}, negative scores are shown in cool (\highd{blue}) colors, while positive scores are shown in warm (\highc{orange}) colors, with rewards normalized to a mean of 0. On the left, we observe that the DPO model is particularly sensitive to errors during chart interpretation from the \chartqa dataset. For instance, when the response incorrectly lists “Lamb” as “Beef” in a chart reading task, the DPO model assigns a highly negative score to this mistake.

On the right, we present examples from the \aitd dataset. Here, a hallucination in the response, such as incorrectly stating that the left side of the moon is illuminated (the correct answer is the right side), receives a low score. Additionally, when external knowledge is required to correctly identify the moon’s phase as “Crescent” instead of “Gibbous,” the DPO model penalizes the incorrect “Gibbous” answer with a negative score.
This indicates that the DPO model is more sensitive to knowledge-based errors than the SFT model, explaining its superior performance on CoT reasoning tasks in datasets such as \aitd.



\section{Conclusion}
In this work, we aim to improve VLM CoT reasoning. First, we collect a CoT reasoning dataset \sharegpto across a broad range of VQA tasks. We demonstrate that fine-tuning on this dataset significantly enhances reasoning performance. Additionally, we further improve these models using reinforcement learning with direct preference optimization, which strengthens their ability to reason and generalize to direct answer prediction tasks. Our results show that these approaches effectively enhance the reasoning capabilities of VLMs, paving the way for more robust and interpretable multimodal models.

\bibliography{iclr2025_conference}
\bibliographystyle{iclr2025_conference}

\clearpage
\appendix
\counterwithin{figure}{section}
\counterwithin{table}{section}

\begin{center}
\Large
\textsc{Content of Appendix}
\end{center}

In this paper, we aim to enhance chain-of-thought (CoT) reasoning in visual language models. In the main paper, we have discussed the CoT data distillation, supervised-finetuning (SFT) and reinforcement learning (Rl) with direct preference optimization (DPO) algorithm. In the appendix, we provide additional items that offer further insight into each aspect:
\begin{itemize}
\item[\ref{appendix:gpto_distill}] \sharegpto Data for VLM CoT Reasoning;
\item[\ref{appendix:gpto_eval}] \gpto Evaluation and Prompt Optimization;
\item[\ref{appendix:baseline_eval}] Baseline Evaluation;
\item[\ref{appendix:dpo_zeroshot}] Nearly Zero Data Learning for CoT Reasoning;
\item[\ref{appendix:sft_experiments}] More SFT Ablation Experiments;
\item[\ref{appendix:dpo_experiments}] More DPO Experiments;

\end{itemize}

\newpage
\section{\sharegpto Data for VLM CoT Reasoning}
\label{appendix:gpto_distill}

\subsection{Prompt for \gpto Distillation}
\Cref{fig:gpt4o_distillation_system} and \cref{fig:gpt4o_distillation} illustrate the \gpto system (task) prompt and the \gpto distillation prompt. We employ the same prompt across all VQA datasets for data distillation. Specifically, the input to the prompt consists of an image, a question, and a short answer. The short answer serves as a reference for \gpto to generate a CoT reasoning followed by a final answer after '\#\#\# Answer'. We show a few more examples in the next subsections.

\begin{figure}[ht]
\centering
\small
\centering
\begin{minted}[fontsize=\footnotesize, frame=single,linenos=false,breaklines,breaksymbol=,escapeinside=||,bgcolor=Box2Color]{text}

When provided with an image, a question, and a reference answer, generate a chain-of-thought step that helps derive your own answer.
Your rationale should include detailed visual elements in order to derive the answer.

\end{minted}
\caption{\gpto system prompt for CoT distillation.}
\label{fig:gpt4o_distillation_system}
\end{figure}

\begin{figure}[ht]
\centering
\small
\centering
\begin{minted}[fontsize=\footnotesize, frame=single,linenos=false,breaklines,breaksymbol=,escapeinside=||,bgcolor=Box1Color]{text}
# Objective #
You are provided with an image, a question and a reference answer. Your job is to generate a rationale that logically derives the answer from the visual clues. 

#############

# Question #
{question}

#############

# Reference Answer #
{answer}

#############

# Rationale Requirement #
1. Do not state an answer at the beginning. Explain the visual clues that help to derive the answer.
2. Don't state that the reference answer is correct or consistent to your finding. Your are writing your own solution.
3. State your own derivation at the end with new line: ### Answer: <your answer>

#############
\end{minted}
\caption{\gpto prompt for CoT distillation.}
\label{fig:gpt4o_distillation}
\end{figure}

\newpage
\subsection{Filtering Mismatched Annotations in Distillation}

\begin{figure}[ht]
    \centering
    \includegraphics[width=\linewidth]{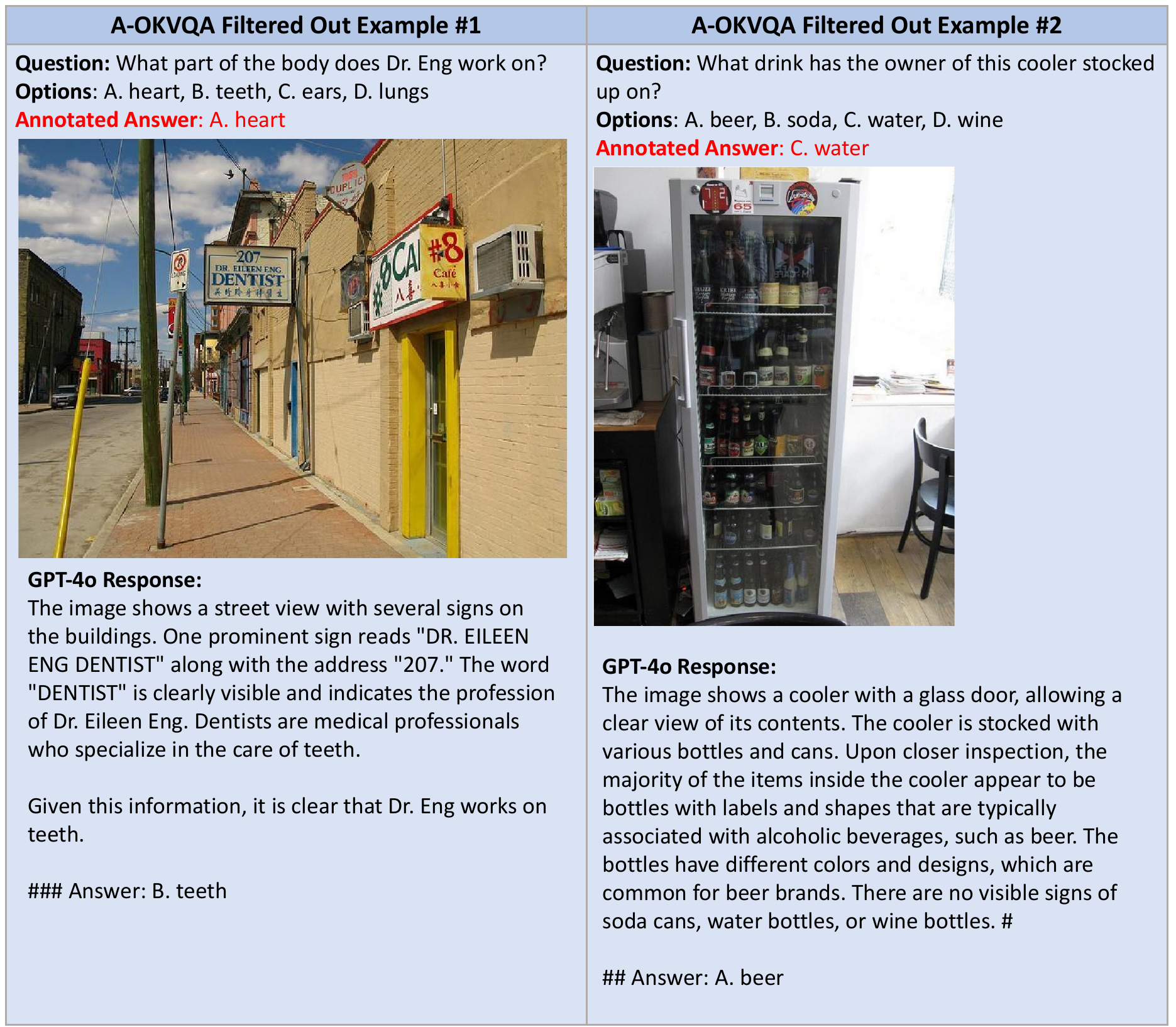}
    \caption{An example from the \aokvqa dataset highlights cases where the annotated answer does not match the \gpto-generated answer. In these cases, the \gpto answers are correct, while the annotations contain labeling errors. In the left figure, the sign reads “dentist” (correctly identified by \gpto), and the answer should relate to ‘teeth,’ not ‘heart’ as in the annotation. In the right figure, the fridge contains beer, but the annotation incorrectly labels it as ‘water.’ Consequently, we filter out instances where the \gpto-generated answer does not match the annotated answers.
}
    \label{fig_apd:filtered_aokvqa}
\end{figure}

\begin{figure}[ht]
    \centering
    \includegraphics[width=\linewidth]{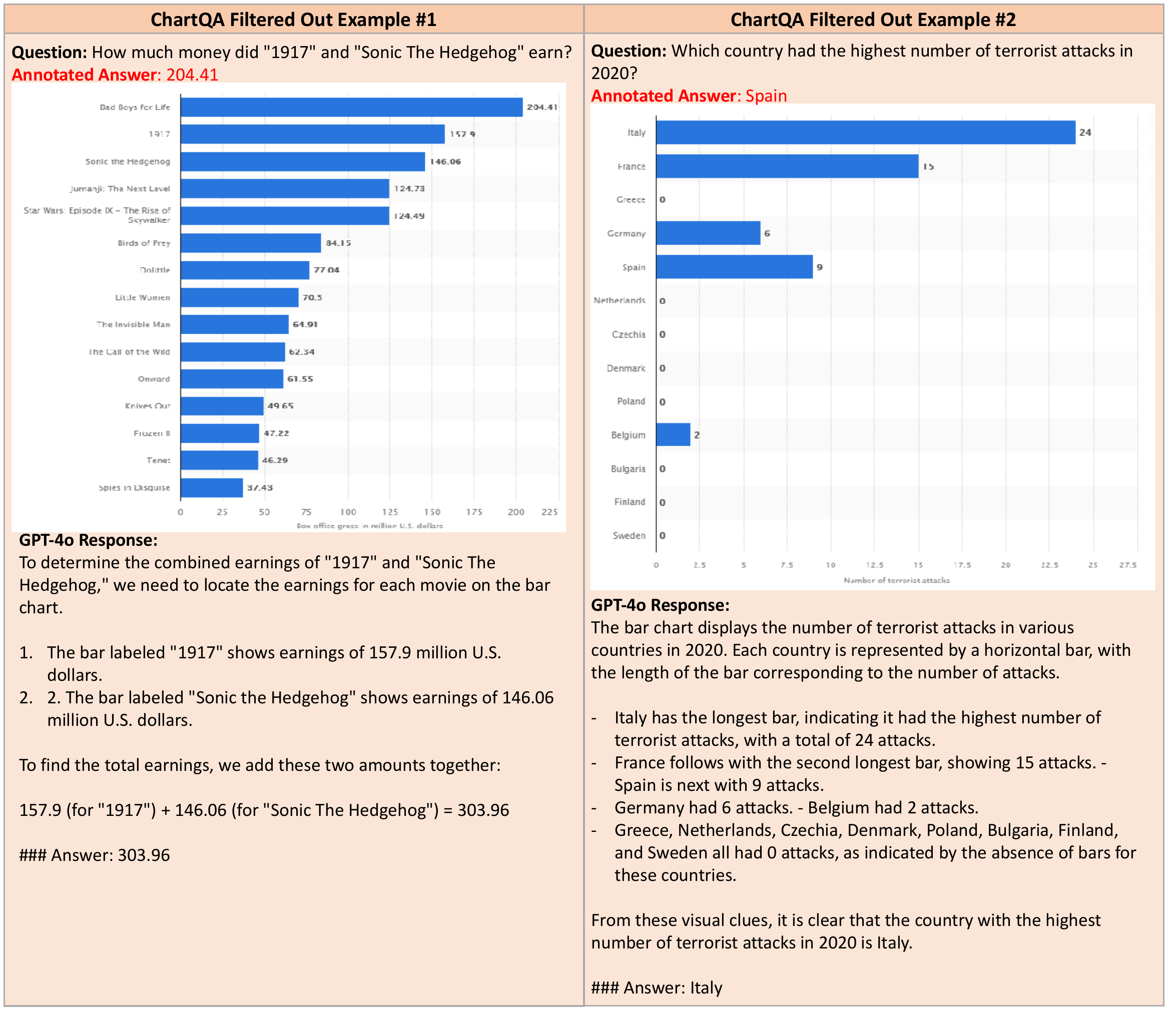}
    \caption{Filtered examples from the \chartqa dataset are shown. In the left figure, \gpto correctly identifies ‘1917’ and ‘Sonic The Hedgehog’ and provides the correct summation, while the annotated answer incorrectly lists ‘204.41’, which is the value for 'Bad Boys for Life' and is unrelated to the question. In the right figure, \gpto accurately ranks the numbers from highest to lowest, but the annotated answer incorrectly identifies ‘Spain’ as having the highest value, when it should be the third largest.
}
    \label{fig_apd:filtered_chartqa}
\end{figure}

In the \gpto prompt shown in \cref{fig:gpt4o_distillation}, we treat the annotation as a \textit{reference answer} and instruct \gpto to generate its own solution based on that reference. In \cref{fig_apd:filtered_aokvqa} and \cref{fig_apd:filtered_chartqa}, we illustrate cases where the \gpto-generated solution differs from the annotated answer. Upon human examination, we identified errors in the annotations. For example, in \cref{fig_apd:filtered_aokvqa}, there are issues such as incorrect text recognition (e.g., “dentist” misidentified as “heart”) and incorrect object identification (e.g., “beer” as “water”). In \cref{fig_apd:filtered_chartqa}, the annotation errors involve incorrect calculations in the left figure and miscounting in the right figure.

To ensure consistency and avoid potential errors, we filtered out examples where the \gpto generated answer differs from the annotated answer. In \sharegpto, we release the SFT CoT data along with the original distillation and filtered examples for reference.

\newpage
\newpage

\newcommand{\aokvqaPromptDirect}{
\texttt{%
Answer the question. Do not write a full sentence, just provide a letter choice.\newline
{question}\newline
\{Question\}
}
}

\newcommand{\chartqaPromptDirect}{
\texttt{%
Answer the question with following instruction:\newline
1. Do not write a full sentence, just provide a value.\newline
2. Don't include any unit, i.e. 56 instead of 56 meters\newline
3. Don't include '\%' sign, i.e. 56 instead of 56\%\newline
\newline
Question: \{Question\}
}
}

\newcommand{\docvqaPromptDirect}{
\texttt{%
Answer the question. Do not write a full sentence, just provide a value.\newline
\newline
Question: \{question\}
}
}

\newcommand{\mathvistaPromptDirect}{
\texttt{%
Answer the question. Do not write a full sentence, just provide a value or letter choice.\newline
\{question\}
}
}

\newcommand{\systemPromptCot}{
\texttt{%
When provided with an image and a question, generate a rationale first and then derive an answer.\newline
Your rationale should include detailed visual elements in order to derive the answer.
}
}

\newcommand{\aokvqaPromptCot}{
\texttt{%
Answer the question with following instruction:\newline
1. Generate a rationale first and then derive an answer.\newline
2. For your final answer, provide a letter choice.\newline
\newline
Question: \newline
\{question\}\newline
\newline
\# Output Format \#\newline
<rationale>\newline
\#\#\# Answer: <your answer>\newline
}
}

\newcommand{\chartqaPromptCot}{
\texttt{%
Answer the question with following instruction:\newline
1. Generate a rationale first and then derive an answer. \newline
2. Don't include any unit, i.e. 56 instead of 56 meters\newline
3. Don't include '\%' sign, i.e. 56 instead of 56\%\newline
\newline
Question: \newline
\{question\}\newline
\newline
\# Output Format \#\newline
<rationale>\newline
\#\#\# Answer: <your answer>\newline
}
}

\newcommand{\docvqaPromptCot}{
\texttt{%
\# Objective \#\newline
You are provided with an image, a question. Your job is to generate a rationale first and then derive an answer.\newline
\newline
\#\#\#\#\#\#\#\#\#\#\#\newline
\newline
\# Question \#\newline
\{question\}\newline
\newline
\#\#\#\#\#\#\#\#\#\#\#\newline
\newline
\# Rationale Requirement \#\newline
1. Do not state an answer at the beginning. Explain descriptions of visual clue that help to derive the answer.\newline
2. Conclude with \#\#\# Answer: <your answer>\newline
3. Your final answer should be a single word or phrase.\newline
4. If possible, copy the answer from document. Don't add or remove symbols, units, or titles.\newline
\newline
\#\#\#\#\#\#\#\#\#\#\#\newline
\newline
\# Output Style \#\newline
<rationale>\newline
\#\#\# Answer: <your answer>\newline
\newline
\#\#\#\#\#\#\#\#\#\#\#
}
}

\newcommand{\textvqaPromptCot}{
\texttt{%
\# Objective \#\newline
You are provided with an image, a question. Your job is to generate a rationale first and then derive an answer.\newline
\newline
\#\#\#\#\#\#\#\#\#\#\#\newline
\newline
\# Question \#\newline
\{question\}\newline
\newline
\#\#\#\#\#\#\#\#\#\#\#\newline
\newline
\# Rationale Requirement \#\newline
1. Do not state an answer at the beginning. Explain descriptions of visual clue that help to derive the answer.\newline
2. Conclude with \#\#\# Answer: <your answer>\newline
3. Your final answer should be a single word or phrase.\newline
4. Output your answer in lower case.\newline
\newline
\#\#\#\#\#\#\#\#\#\#\#\newline
\newline
\# Output Style \#\newline
<rationale>\newline
\#\#\# Answer: <your answer>\newline
\newline
\#\#\#\#\#\#\#\#\#\#\#
}
}

\newcommand{\ocrbenchPromptCot}{
\texttt{%
Answer the question with following instruction:\newline
1. Generate a rationale first and then derive an answer.\newline
2. Your answer should be a single word or phrase.\newline
\newline
Question: \newline
\{question\}\newline
\newline
\# Output Format \#\newline
<rationale>\newline
\#\#\# Answer: <your answer>\newline
}
}

\newcommand{\mathvistaPromptCot}{
\texttt{%
\# Objective \#\newline
You are provided with an image, a question. Your job is to generate a rationale that logically derives an answer from the visual clues. \newline
\newline
\#\#\#\#\#\#\#\#\#\#\#\newline
\newline
\# Question \#\newline
\{question\}\newline
\newline
\#\#\#\#\#\#\#\#\#\#\#\newline
\newline
\# Rationale Requirement \#\newline
1. Do not state an answer at the beginning. Explain step by step logic to derive the answer.\newline
2. Conclude with \#\#\# Answer: <your answer>\newline
\newline
\#\#\#\#\#\#\#\#\#\#\#\newline
\newline
Example output style:\newline
\newline
<rationale>\newline
\#\#\# Answer: <your answer>\newline
\newline
\#\#\#\#\#\#\#\#\#\#\#
}
}

\section{\gpto Evaluation and Prompt Optimization}
\label{appendix:gpto_eval}
In this section, we present the prompts used for \gpto on benchmark datasets, including both direct and Chain-of-Thought (CoT) predictions. Similar to the findings in \cite{borchmann2024notes}, we observed that \gpto's performance is highly sensitive to prompt phrasing. We explored several sets of prompts and selected the best-performing ones for reporting results. Specifically, we try to align our results with those reported in \cite{onevision, tong2024cambrian}, Claude 3.5 Sonnet for Vision ~\footnote{https://www.anthropic.com/news/claude-3-5-sonnet}, among others.

\paragraph{Prompt Optimization}
We follow the process outlined in \cite{borchmann2024notes} to design effective \gpto prompts for the benchmark datasets. A random subset of 200 instances is selected as a development set to evaluate manually designed prompts. We manually inspect the predicted results and identify issues such as the model being overly cautious in declining answers, incorrect output formatting, or style mismatches with the ground truth labels. As an illustrative example, we detail the prompt optimization process using \chartqa, and apply similar techniques to the other datasets. Finally, we provide the prompts used for replicating our test results.

\begin{table}[ht]
\setlength{\extrarowheight}{3pt}
\centering
\caption{Prompt optimization on \chartqa for \textbf{direct} prediction evaluated with relaxed accuracy.\label{tab:gpto_direct_prompt_chartqa}}
\begin{tabular}{c p{10cm} >{\centering\arraybackslash}m{2cm}}
\hline
\# & \textbf{Prompt} & \textbf{\chartqa (relaxed acc)}  \\ \toprule
1 & \texttt{\{Question\}} & 2.7 \\ 
2 & \texttt{\{Question\}\newline Answer the question directly.} & 32.3 \\ 
3 & \texttt{Answer the question. Do not write a full sentence, just provide a value.\newline Question: \{Question\}} & 56.4 \\
4 & \texttt{Answer the question with following instruction:\newline
1. Do not write a full sentence, just provide a value.\newline
2. Don't include any unit, i.e. 56 instead of 56 meters\newline
Question: \{Question\}}
& 75.2 \\
5 & \chartqaPromptDirect
& \bf 80.3 \\
\bottomrule
\end{tabular}
\end{table}

We apply the prompts described in \cref{tab:gpto_direct_prompt_chartqa} to the development set and compare the predictions with the ground truth to optimize the prompts. Specifically, when using prompts \#1 or \#2, \gpto often generates full sentences instead of short answers. While prompt \#3 produces a short answer, it often includes units or special tokens. To address this, we refined the instructions in prompt \#4 by specifying that units should not be included in the final answer. This adjustment improved accuracy from 56.4 to 75.2. We also observed that the ground truth does not contain the \% symbol, which could mismatch in evaluation, and we explicitly include this rule in prompt \#5. Finally, we applied the tuned prompt to the test set, achieving an accuracy of $79.64$ reported in \cref{tab:sota_compare}.

\newpage

\begin{table}[ht]
\setlength{\extrarowheight}{3pt}
\centering
\caption{Prompt optimization on \chartqa for \textbf{CoT} prediction evaluated with relaxed accuracy.\label{tab:gpto_cot_prompt_chartqa}}
\begin{tabular}{c p{10cm} >{\centering\arraybackslash}m{2cm}}
\hline
 & \textbf{System Prompt} & \textbf{\chartqa (relaxed acc)}  \\ \toprule
 & \systemPromptCot &  \\
\# & \textbf{Prompt} &  \\ \hline
1 & 
\chartqaPromptCot
& \bf 84.7 \\
2 & Prompt \#1, removing system prompt & 84.1 \\
\bottomrule
\end{tabular}
\end{table}

In \cref{tab:gpto_cot_prompt_chartqa}, we first introduce output format instructions to guide \gpto in generating the correct CoT format, which aids in extracting the final answer. We reused the criteria from the direct prediction prompt to evaluate the results. Additionally, we found that including a system prompt typically leads to a 0.5-point increase in score across datasets, although it does not improve direct answer prediction. We hypothesize that the system prompt helps \gpto adhere more closely to the CoT output format. Finally, we applied the tuned prompt to the test set, achieving an accuracy of $84.72$ reported in \cref{tab:sota_compare}. 

Following the prompt optimization steps outlined above, we provide the prompts used to replicate our \gpto test results in the next section.

\newpage

\subsection{\gpto Prompts for Evaluation}
\Cref{tab:gpto_direct_prompts} and \cref{tab:gpto_cot_prompts} provide the optimized prompts for benchmark dataset evaluation. The tuning process does not garantee the prompt is optimal, but that roughly matches the reported value from previous papers \cite{onevision, tong2024cambrian}, Claude 3.5 Sonnet for Vision ~\footnote{https://www.anthropic.com/news/claude-3-5-sonnet}, among others. We include the prompts for reference to replicate the \gpto results on benchmark datatsets.

\begin{table}[ht]
\setlength{\extrarowheight}{3pt}
\centering
\caption{Prompts for \textbf{direct} prediction with \gpto on benchmark datasets.\label{tab:gpto_direct_prompts}}
\begin{tabular}{p{2cm} p{11cm} }
\hline
Dataset & \textbf{Prompt}   \\ \toprule
\aokvqa \newline \aitd \newline \sqa \newline \mmstar  & \aokvqaPromptDirect \\
\hline
\chartqa & \chartqaPromptDirect  \\
\hline
\docvqa \newline \textvqa \newline \infovqa \newline \ocrbench & \docvqaPromptDirect \\
\hline
\mathvista \newline \mmmu & \mathvistaPromptDirect \\
\bottomrule
\end{tabular}
\end{table}

\newpage
\begin{longtable}{p{2cm} p{11cm}}
\caption{Prompts for \textbf{CoT} prediction with \gpto on benchmark datasets.\label{tab:gpto_cot_prompts}} \\
\hline
Dataset & \textbf{CoT Prompt}   \\ \toprule
\endfirsthead
\multicolumn{2}{c}{{\tablename\ \thetable{} -- continued from previous page}} \\
\hline
Dataset & \textbf{Prompt}   \\ \toprule
\endhead
\hline \multicolumn{2}{r}{{Continued on next page}} \\
\endfoot
\bottomrule
\endlastfoot
\bf system\newline prompt & \systemPromptCot \\
\hline
\aokvqa \newline \aitd \newline \sqa \newline \mmstar & \aokvqaPromptCot \\
\hline
\chartqa & \chartqaPromptCot  \\
\hline
\docvqa \newline \infovqa & \docvqaPromptCot \\
\hline
\textvqa & \textvqaPromptCot \\
\hline
\ocrbench & \ocrbenchPromptCot \\
\hline 
\mathvista \newline \mmmu & \mathvistaPromptCot \\
\end{longtable}


\newpage

\section{Baseline Evaluation}
\label{appendix:baseline_eval}

\begin{table}[ht]
\caption{
Evaluation of VLM performance on benchmark datasets with direct and CoT inference. \label{tab_apd:baseline_eval}
}
\centering
\begin{tabular}{l|cc|cc}
    \toprule
    \textbf{Dataset} & \multicolumn{2}{c}{\modelbase} & \multicolumn{2}{c}{\modelformat}  \\
    \cmidrule(lr){2-3} \cmidrule(lr){4-5}
    & \textbf{direct} & \textbf{CoT} & \textbf{direct} & \textbf{CoT}\\
    \midrule
    \aok & 85.9 & 44.5 & 85.8 & 84.3  \\
    \chartqa & 68.6 & 52.8 & 70.2 & 71.2 \\
    \docvqa & 78.4 & 57.1 & 75.7 & 67.0   \\
    \infovqa & 36.6 & 25.8 & 37.7 & 34.9  \\
    \textvqa & 67.2 & 41.6 & 68.2 & 62.2   \\
    \aitd & 73.0 & 70.0 & 71.5 & 67.4  \\
    \sqa & 77.4 & 75.8 & 75.4 & 74.4   \\
    \mathvista & 37.3 & 25.3 & 39.3 & 40.3  \\
    \ocrbench & 57.7 & 59.7 & 59.1 &  56.6  \\
    \mmstar & 47.8 & 45.7 & 44.7 & 46.7  \\
    \mmmu & 42.8 & 37.6 & 41.8 & 37.7   \\
    \hline  
    Avg & 61.2 & 48.7 & 60.9 & 58.4 \\
    \bottomrule
\end{tabular}
\end{table}

In this section, we provide evaluation details for our base model, which uses the \llavanext architecture with weights initialized from \openllavanext. We selected \openllavanext weights because the data and training pipelines were fully available at the time of model development, allowing us to avoid reliance on the unreleased real user interactions referenced in \cite{liu2024llavanext}. The pretraining data for \openllavanext consists of 1M image-text pairs, sourced from datasets such as ShareGPT4V, ALLaVA-Instruct-VFLAN-4V, DocVQA, SynDog-EN, ChartQA, DVQA, AI2D, and GeoQA+. 

\begin{figure}
    \centering
    \includegraphics[width=1\linewidth]{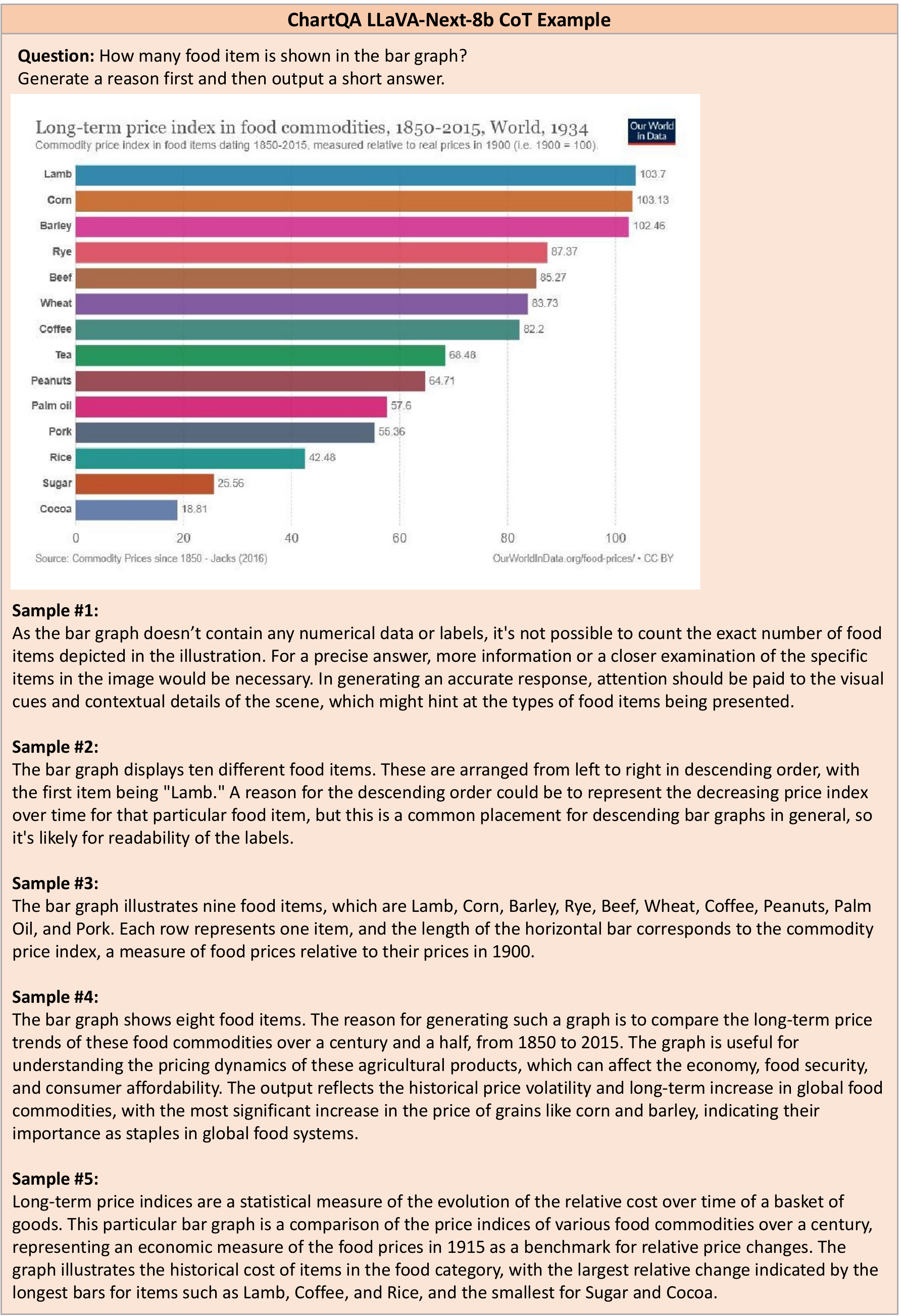}
    \caption{Randomly sampled examples from \modelbase with temperature=1.0 for a test case in \chartqa reveal that the model struggles to effectively follow the CoT prompt. In Sample 1, the model refuses to answer the question. In Samples 2-4, the model generates an answer first, followed by an explanation. In the final sample, the model produces a description instead of reasoning through the question, without providing an answer.}
    \label{fig_apd:modelbase_example}
\end{figure}

When evaluating \modelbase, we identified several issues, such as the inability to follow the CoT prompt, refusal to answer questions, and generating irrelevant reasoning. In \cref{fig_apd:modelbase_example}, we present randomly sampled examples from \modelbase with a temperature setting of 1.0 on a \chartqa test case. These examples demonstrate the model’s difficulty in adhering to the CoT prompt. In the first example, the model declines to answer the question. In the second to fourth examples, the model provides an answer first, followed by an explanation, which doesn't effectively use thought process to answer the question. In the final example, the model generates a descriptive response instead of reasoning through the question, ultimately failing to provide an answer. This illustrates the model’s inconsistent handling of the prompt structure.

\Cref{tab_apd:baseline_eval} presents the evaluation results for \modelbase. For CoT predictions, we use \chatgpt to extract a letter choice or short answer from the long-form model output, using the prompts shown in \cref{fig:chatgpt_mcq_prompt} and \cref{fig:chatgpt_sa_prompt}. However, due to \modelbase’s inability to accurately follow the CoT format, its performance is significantly worse compared to direct predictions and our format-aligned model. For direct prediction, our \modelformat has similar performance as that of \modelbase.

\begin{figure}
    \centering
    \includegraphics[width=1\linewidth]{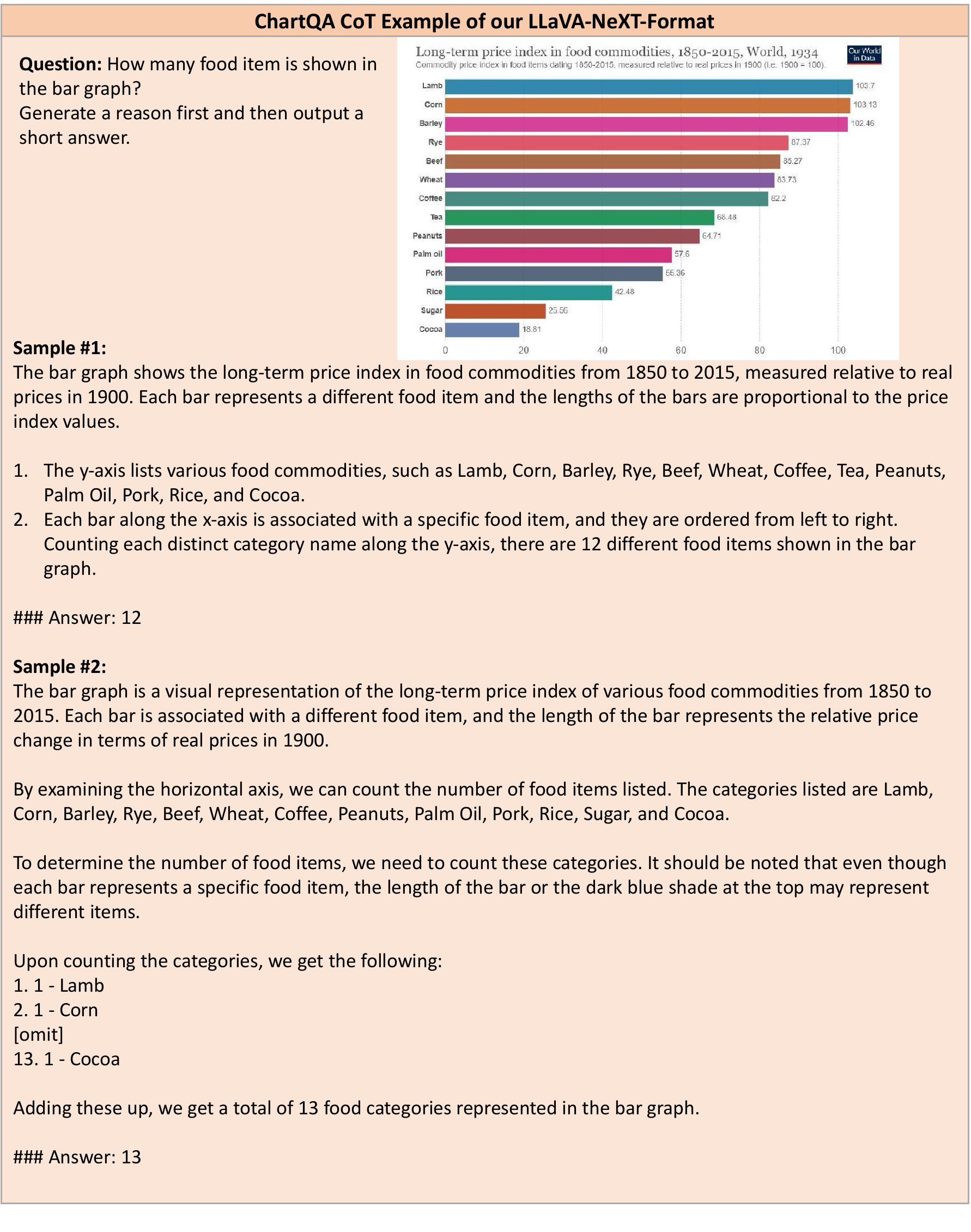}
    \caption{Randomly sampled examples from \modelformat with a temperature setting of 1.0, evaluated on the same test case in \chartqa, show that after training on 450 format-aligned data, the model is able to follow the CoT prompt by verbalizing the thought process and providing a short answer.}
    \label{fig_apd:modelformat_example}
\end{figure}

In \cref{fig_apd:modelformat_example}, we present the same example trained with our format-aligned data for CoT using only 450 examples. The model successfully follows the CoT format by verbalizing the thought process and providing a short answer after "\#\#\# Answer:". This allows us to use a rule-based extractor to retrieve answers, which also improves CoT performance, as shown in \cref{tab_apd:baseline_eval}. However, the example also demonstrates that, while our data induces the CoT process, the reasoning remains incorrect. Sampling 32 examples using the format in \cref{fig_apd:modelformat_example} resulted in 3 correct answers, yielding an accuracy of 9.4\% for that case. As a result, only slight gain is observed on \chartqa for CoT prediction vs. direct prediction.

Based on the above experiments, we report \modelbase as the baseline in the paper to avoid the difficulty in CoT evaluation and answer extraction of \llavanext baseline.

\begin{figure}
\centering
\small
\centering
\begin{minted}[fontsize=\footnotesize, frame=single,linenos=false,breaklines,breaksymbol=,escapeinside=||,bgcolor=Box1Color]{text}

You are an AI assistant who will help me to match an answer with several options of a single-choice question. You are provided with a question, several options, and an answer, and you need to find which option is most similar to the answer. If the meaning of all options are significantly different from the answer, output Z. Your should output a single uppercase character in A, B, C, D (if they are valid options), and Z. 
Example 1: 
Question: What is the main object in image?
Options: A. teddy bear B. rabbit C. cat D. dog
Answer: a cute teddy bear
Your output: A
Example 2: 
Question: What is the main object in image?
Options: A. teddy bear B. rabbit C. cat D. dog
Answer: Spider
Your output: Z
Example 3: 
Question: {question}
Options: {options}
Answer: {answer}
Your output: 

\end{minted}
\caption{\chatgpt answer extraction prompt for multiple-choices questions.}
\label{fig:chatgpt_mcq_prompt}
\end{figure}

\begin{figure}
\centering
\small
\centering
\begin{minted}[fontsize=\footnotesize, frame=single,linenos=false,breaklines,breaksymbol=,escapeinside=||,bgcolor=Box1Color]{text}

Your goal is to extract a short answer from a chain-of-thought prediction. You are given a question and model prediction, the image is omitted. 
You need to determine the answer from the prediction. If no answer can be derive, output NA.

###### Example 1 ######
### Question: 
How many bars are there in the chart?
### Prediction: 
The result shows bar graphs ..., counting the bars, there are a total of 8 bars.
### Your output: 
8

###### Example 2 ######
### Question: 
Determine the date appeared in the document.
### Prediction: 
The figure displays a document on financial income ... the date 2008/01/15 appears at body of text.
### Your output: 
2008/01/15

###### Your Task ######
### Question: 
{question}
### Prediction: 
{prediction}
### Your output: 

\end{minted}
\caption{\chatgpt answer extraction prompt for short answer questions.}
\label{fig:chatgpt_sa_prompt}
\end{figure}

\newpage
\section{Nearly Zero Data Learning for CoT Reasoning}
\label{appendix:dpo_zeroshot}

\begin{table}[ht!]
\centering
\caption{
We study a self-taught reasoner with minimal CoT data (only 450 format-aligned examples). \modeldirect is used as the baseline, and our LLaVA-Next-STaR is trained with a rejection sampling method. The best CoT predictions are highlighted in \highc{orange}, and the best direct predictions are highlighted in \highd{blue}. Our rejection sampling method outperforms both CoT and direct prediction, with the exception of two data points.
}
\resizebox{1.0\linewidth}{!}{
    \begin{tabular}{lccccccccc}
    \toprule
    \textbf{Methods}
    & \textbf{Prompting}
    & \aok
    & \chartqa
    & \docvqa 
    & \infovqa
    & \textvqa
    & \aitd
    & \sqa
    & \mathvista
    \\
    \midrule

LLaVA-Next & direct & \highd{86.4}	& 73.7	& 78 & 45.4 & 71.9 & 78.8 & 91.5 & 43.2  \\
+ Direct \cTwo & CoT & 85.7 & 71.8 & 68.8 & 38.6 & \highc{63.6} & 72.5 & 85.4 & 38.6  \\

\hline
LLaVA-Next & direct & 85.9 & \highd{74.6} & \highd{79.2}	& \highd{47.4}	& \highd{72.1}	& \highd{79.5}	& \highd{92.2}	& \highd{44.4}  \\
-STaR & CoT & \highc{85.9} & \highc{77.9}	& \highc{75.8}	& \highc{44.0} & 25.1	& \highc{76.6}	& \highc{86.8}	& \highc{42.0} \\

\bottomrule
\hline
\end{tabular}
}

\label{tab_apd:rl_zeroshot}
\end{table}

In this section, we demonstrate how minimal CoT training data can enhance CoT reasoning capabilities. Specifically, we use only 450 CoT format-aligned examples alongside all available direct prediction data, with \modeldirect as the baseline. We apply rejection sampling fine-tuning (RFT) following~\citep{sun2024easy, setlur2024rl} to train a self-taught chain-of-thought reasoner, denoted as LLaVA-Next-STaR. From \modeldirect, we sample 32 CoT examples for each training instance and select those whose final predictions match the ground truth. Up to three positive examples are selected per question, resulting in a dataset of 260k RFT examples.

As shown in \cref{tab_apd:rl_zeroshot}, RFT training improves both CoT reasoning and direct predictions overall, with the exception of two data points. Notably, \textvqa shows a significant drop in CoT performance, which we will explore further in future work. Notable (>3\%) gain is observed on \chartqa, \docvqa, \infovqa, \aitd and \mathvista, and roughly 1\% gain is observed on direct prediction on those datasets as well. 

\paragraph{DPO Experiments} Prior to the RFT experiments, we conducted DPO experiments on the \chartqa dataset under the same conditions as described in \cref{sec:sft}. However, the improvements were modest, with a 72.3 (+0.5) gain in CoT prediction and a 74.2 (+0.5) gain in direct prediction. In contrast, RFT yielded a significant improvement, with 77.9 (+6.1) on CoT prediction and 74.6 (+0.9) on direct prediction. We hypothesize that for models with relatively weak CoT reasoning capabilities, RFT may be more effective in enhancing model performance, whereas DPO with preference modeling may be less impactful. We leave further analysis for future work.

\newpage
\section{SFT Ablation Experiments}
\label{appendix:sft_experiments}

{\small
\begin{longtable}{lccccccccc}
\caption{SFT Ablation Results: For each dataset, ‘-C’ indicates the inclusion of CoT data for training, and ‘-D’ indicates the inclusion of direct prediction data.}\label{tab_apd:sft_ablation}\\
\toprule
\textbf{Methods} & \textbf{Prompt} & \aok & \chartqa & \docvqa & \infovqa & \textvqa & \aitd & \sqa & \mathvista \\
\midrule
\endfirsthead

\toprule
\textbf{Methods} & \textbf{Prompting} & \aok & \chartqa & \docvqa & \infovqa & \textvqa & \aitd & \sqa & \mathvista \\
\midrule
\endhead

\midrule
\multicolumn{10}{r}{\textit{Continued on next page}} \\
\midrule
\endfoot

\bottomrule
\endlastfoot

\multirow{2}{*}{\modelformat} & direct & 85.8 & 70.2 & 75.7 & 37.7 & 68.2 & 71.5 & 75.4 & 39.3 \\ 
 & cot & 84.3 & 71.2 & 67 & 34.9 & 62.2 & 67.4 & 74.4 & 40.3 \\ 
\hline

\multirow{2}{*}{\chartqa-C+D} & direct & 85 & 74.9 & 75.8 & 36.5 & 68.2 & 72.2 & 77.4 & 42.8 \\ 
 & cot & 84.4 & 81.7 & 69 & 32.2 & 63.3 & 68.6 & 74.9 & 41.7 \\ 
\multirow{2}{*}{\chartqa-D} & direct & 85.2 & 73.1 & 74.6 & 34.1 & 67.1 & 71.5 & 76.4 & 40.3 \\ 
 & cot & 84.3 & 71.8 & 62.4 & 31.8 & 58 & 66.3 & 74 & 35.5 \\ 
\multirow{2}{*}{\chartqa-C} & direct & 85.1 & 70.8 & 74.5 & 35 & 67.9 & 71.6 & 76.9 & 35.3 \\ 
 & cot & 84.9 & 81.4 & 67.2 & 32.2 & 61.5 & 68.8 & 76.6 & 40.1 \\
\hline

\multirow{2}{*}{\aok-C+D} & direct & 86.2 & 69.2 & 75.4 & 37.7 & 67.3 & 70.7 & 77.5 & 38.8 \\ 
 & cot & 84.6 & 70.2 & 67.3 & 36 & 61.6 & 67.2 & 75.8 & 39.8 \\ 
\multirow{2}{*}{\aok-D} & direct & 85.1 & 69 & 75.3 & 38.5 & 66.9 & 72.2 & 76.1 & 39.5 \\ 
 & cot & 84 & 67.7 & 66.5 & 34.8 & 61.1 & 68.4 & 76 & 39.9 \\ 
\multirow{2}{*}{\aok-C} & direct & 84.4 & 69.4 & 75.8 & 37.4 & 67.9 & 69.2 & 77.3 & 34.6 \\ 
 & cot & 84.1 & 69.2 & 67.6 & 35.5 & 59.4 & 67.6 & 74.5 & 40.6 \\ 
\hline

\multirow{2}{*}{\docvqa-C+D} & direct & 85.5 & 69.5 & 80.7 & 40.4 & 68.8 & 72 & 77.5 & 41.1 \\ 
 & cot & 83.9 & 70.9 & 80 & 40.2 & 64.1 & 68.2 & 73.4 & 39.3 \\ 
\multirow{2}{*}{\docvqa-D} & direct & 85.5 & 66.5 & 77 & 39.1 & 68.2 & 70.8 & 76.3 & 41.9 \\ 
 & cot & 83.9 & 66 & 66.4 & 33.7 & 59.9 & 64.8 & 74.5 & 39.3 \\ 
\multirow{2}{*}{\docvqa-C} & direct & 85.2 & 69.1 & 79.1 & 37.5 & 68.5 & 72 & 76.7 & 33.8 \\ 
 & cot & 84.4 & 71.2 & 78 & 38.5 & 63.5 & 68.5 & 74.1 & 38 \\ 
\hline

\multirow{2}{*}{\infovqa-C+D} & direct & 85.8 & 63.4 & 77.1 & 47.7 & 67.6 & 72.5 & 78.1 & 43.6 \\ 
 & cot & 85.3 & 65.4 & 72.6 & 47.5 & 62.4 & 69.4 & 74.6 & 37.8 \\ 
\multirow{2}{*}{\infovqa-D} & direct & 85.7 & 56.7 & 75 & 45.4 & 67 & 72.5 & 77.5 & 42.8 \\ 
 & cot & 83.7 & 53 & 63.5 & 37.8 & 58.2 & 67 & 75 & 37 \\ 
\multirow{2}{*}{\infovqa-C} & direct & 85.2 & 68.3 & 76.5 & 42.5 & 67.8 & 72.5 & 78.2 & 39 \\ 
 & cot & 83.7 & 63.4 & 71.1 & 46.3 & 59.9 & 67.4 & 74.3 & 37.6 \\ 
\hline

\multirow{2}{*}{\textvqa-C+D} & direct & 85.1 & 69.8 & 75.5 & 38.7 & 73 & 71.9 & 76.9 & 42.6 \\ 
 & cot & 84.6 & 68.9 & 70.5 & 36.3 & 70.9 & 67.6 & 76.6 & 36.1 \\ 
\multirow{2}{*}{\textvqa-D} & direct & 84.9 & 68.6 & 74.5 & 37.6 & 71.8 & 70.8 & 77 & 41.7 \\ 
 & cot & 84.4 & 63.3 & 64.2 & 33.2 & 64.2 & 66.1 & 73.6 & 38.2 \\ 
\multirow{2}{*}{\textvqa-C} & direct & 84.6 & 69.1 & 74.6 & 36.9 & 71.4 & 71.9 & 77.1 & 36.6 \\ 
 & cot & 84.7 & 68.2 & 69.5 & 36.9 & 70.3 & 67.8 & 75.1 & 37.1 \\ 
\hline
 
\multirow{2}{*}{\sqa-C+D} & direct & 85.7 & 69 & 75 & 38.4 & 67.3 & 72.3 & 90.2 & 38.7 \\ 
 & cot & 83.1 & 71.2 & 66.5 & 35.6 & 58.9 & 66.9 & 90.4 & 40.8 \\ 
\multirow{2}{*}{\sqa-D} & direct & 84.9 & 68.1 & 74.3 & 37 & 66.8 & 72.2 & 89.2 & 41.3 \\ 
 & cot & 83 & 68.4 & 67.5 & 33.8 & 62.1 & 68.7 & 81.9 & 39.8 \\ 
\multirow{2}{*}{\sqa-C} & direct & 84 & 69.3 & 76 & 38.3 & 68.2 & 71.7 & 85 & 39.2 \\ 
 & cot & 82 & 69 & 65.3 & 34.4 & 58.3 & 66.6 & 88.8 & 39.4 \\ 
 \hline
 
\multirow{2}{*}{\aitd-C+D} & direct & 85.2 & 69.6 & 75.8 & 39 & 67.6 & 78 & 78.4 & 40.1 \\ 
 & cot & 83.8 & 70.2 & 68 & 35.9 & 60.7 & 76.3 & 76.6 & 42.1 \\ 
\multirow{2}{*}{\aitd-D} & direct & 86.3 & 69.2 & 75.1 & 37.3 & 67.2 & 76.8 & 77.6 & 39.7 \\ 
 & cot & 82.7 & 67.6 & 66 & 33.7 & 61.4 & 71.7 & 74.4 & 38.3 \\ 
\multirow{2}{*}{\aitd-C} & direct & 84.4 & 69.6 & 75.9 & 37.7 & 68.2 & 75 & 76.3 & 39.1 \\ 
 & cot & 83.1 & 70.2 & 65.9 & 35.6 & 59.9 & 75.1 & 74.5 & 39.5 \\ 
 \hline
 
\multirow{2}{*}{math-C+D} & direct & 85.3 & 68.5 & 75.5 & 37.8 & 67.3 & 71.7 & 77.4 & 42.7 \\ 
 & cot & 84.4 & 69.7 & 64.3 & 34.2 & 59.3 & 68.7 & 76.3 & 49 \\ 
\multirow{2}{*}{math-C} & direct & 85.2 & 68.1 & 75.6 & 38 & 67.4 & 72 & 77.5 & 40.5 \\ 
 & cot & 84.3 & 70.6 & 66.2 & 34.7 & 59.8 & 68.2 & 78.4 & 45.4 \\ 
\multirow{2}{*}{math+\chartqa} & direct & 85.3 & 70.9 & 75.7 & 36.8 & 67.8 & 71.7 & 78.3 & 41.9 \\ 
 & cot & 84.1 & 81.9 & 67 & 32.6 & 60.7 & 68.3 & 75.5 & 49.7 \\ 
\hline

\multirow{2}{*}{\modelsft} & direct & 85.4 & 76.1 & 82.9 & 50.6 & 73.1 & 79.4 & 90.4 & 44.3 \\
 & cot & 86.2	& 83.0	& 81.8	 &51.6	 & 71.1	 & 78.5	 & 92.7 & 	50.6 \\

\end{longtable}
}

In \cref{tab_apd:sft_ablation}, we present additional ablation experiments on SFT across each dataset, using three settings: direct only, CoT only, and direct + CoT. Additionally, format-aligned data is incorporated during training to enable the model to follow the specific direct or CoT format during inference.
\newpage
\section{Additional DPO Experiments}
\label{appendix:dpo_experiments}

\begin{table}[ht]
    \centering
    \caption{Truncating response length affects the final performance of DPO. No truncation leads to a decline in performance, while truncating to 90 tokens empirically yields the best results.}
    \begin{tabular}{cccccc|c}
    \toprule
        Data/Truncate Len & prompting & 70 & 90 & 110 & No Truncate & SFT baseline \\
        \hline
        \multirow{2}{*}{\chartqa} 
        & direct  & 76.5  & 76.2 & 76.7 & 75.9 & 76.1\\
        & CoT  & 83.9 & 84.2 & 81.8 & 80.6 & 83.0 \\
        \hline
        \multirow{2}{*}{\aokvqa} 
        & direct & 85.2 & 85.2 & 85.3 & 85.1 & 85.4\\
        & CoT & 86.7 & 86.9 & 86.3 & 85.7 & 86.2 \\
    \bottomrule
    \end{tabular}

    \label{tab_apd:dpo_truncate}
\end{table}

\paragraph{Truncating Responses for DPO} In our initial experiments, we observed that truncating response length impacts the final performance of DPO. As shown in \cref{tab_apd:dpo_truncate}, no truncation results in a decline in performance, while truncating to 90 tokens empirically produces the best results. Consequently, we applied a 90-token truncation for the DPO experiments.

\begin{table}[ht]
    \centering
    \caption{Comparison of DPO with the RFT method. The upper part of the table presents the SFT baseline and the DPO model, while the lower part shows the ablation results of RFT trained on each of the \aok, \chartqa, and math training datasets, as well as their combined results.}
    \begin{tabular}{ccccc}
    \toprule
        Methods & prompting & \aok & \chartqa & \mathvista \\
        \hline
        \multirow{2}{*}{SFT baseline} 
        & direct & 85.4 & 76.1 & 44.3  \\
        & CoT & 86.2 & 83.0  & 50.6 \\
        \hline
        \multirow{2}{*}{\modelrl} 
        & direct & 85.4 & 76.4 & 44.2  \\
        & CoT & 87.0 & 84.2  & \bf 52.1 \\
        \hline\hline
        \aokvqa & direct  & 85.1  & 72.7 & 37.4  \\
        -RFT & CoT  & \bf 87.7 & 0.0 & 32.5  \\
        \hline
        \aokvqa & direct  & 85.8  & 74.9 & 41.3  \\
        -RFT+Format & CoT  & 86.3 & 80.2 & 46.5  \\
        \Xhline{1pt}
        \chartqa & direct  & 85.4  & 75.0 & 42.6  \\
        -RFT & CoT  & 86.7 & 83.9 & 52.0  \\
        \hline
        \chartqa & direct  & 85.9  & 75.8 & 44.4  \\
        -RFT+Format & CoT  & 85.5 & 83.4 & 50.6  \\
        
        \Xhline{1pt}
        Math & direct  & 85.3  & 76.0 & 32.4  \\
        -RFT & CoT  & 86.7 & 67.3 & 50.9  \\
        \hline
        Math & direct  & 85.5  & 76.0 & 39.6  \\
        -RFT+Format & CoT  & 85.5 & 82.0 & 50.0  \\

        \Xhline{1pt}
        Combined & direct  & 85.3  & 75.4 & 37.8  \\
        -RFT & CoT  & 85.4 & \bf 84.4 & 49.0  \\
        \hline
        Combined & direct  & 85.0  & 75.5 & 43.0  \\
        -RFT+Format & CoT  & 86.6 & 83.1 & 47.1  \\

    \bottomrule
    \end{tabular}

    \label{tab_apd:dpo_rft}
\end{table}

\paragraph{DPO vs. RFT} Following \cref{appendix:dpo_zeroshot}, we examine the impact of RFT and compare it to the DPO method.

In \cref{tab_apd:dpo_rft}, for \aokvqa, we observe that training with A-OKVQA RFT alone yields the best result for \aokvqa; however, the model’s ability to generate short answers is entirely lost. When format-aligned data is added, there is a trade-off between performance on \aokvqa and other datasets.

When the datasets are combined for training, we see improvements only on \chartqa, while performance on \aokvqa and \mathvista declines. This indicates that balancing RFT across datasets is challenging, especially when the SFT model already performs relatively well on basic tasks. In contrast, the DPO model demonstrates consistent gains across datasets, showing better generalization.

\end{document}